\pdfoutput=1
\documentclass[sigconf, nonacm]{acmart}

\setcopyright{none}
\renewcommand\footnotetextcopyrightpermission[1]{}

\usepackage{amsmath}
\usepackage{colortbl}
\usepackage{multirow}
\usepackage{tabularx}
\newcolumntype{Y}{>{\raggedright\arraybackslash}X}
\usepackage{enumitem}
\usepackage{xspace}
\usepackage{tikz}
\usetikzlibrary{arrows.meta,calc,positioning,shapes.geometric,patterns}
\usepackage{pgfplots}
\pgfplotsset{compat=1.18}
\usepgfplotslibrary{groupplots}
\usepackage[most]{tcolorbox}
\usepackage{pifont}
\usepackage{subcaption}

\newcommand{\Omc}{\ensuremath{\mathcal{O}}\xspace}

\newcommand{\ELp}{\ensuremath{\mathcal{EL}^{+}}\xspace}
\newcommand{\NC}{\ensuremath{\mathsf{N_C}}\xspace}
\newcommand{\NR}{\ensuremath{\mathsf{N_R}}\xspace}

\newcommand{\oursname}{OA-RAG}
\newcommand{\ours}{\oursname\xspace}
\newcommand{\code}[1]{\texttt{#1}}

\newif\ifshowrevisions
\showrevisionstrue
\definecolor{revblue}{RGB}{0,84,200}
\definecolor{suggorange}{RGB}{214,96,0}
\ifshowrevisions
  \colorlet{revtext}{revblue}
  \colorlet{fixtext}{red}
  \colorlet{slotcolor}{teal!75!black}
\else
  \colorlet{revtext}{black}
  \colorlet{fixtext}{black}
  \colorlet{slotcolor}{blue}
\fi
\newcommand{\rev}[1]{\ifvmode\leavevmode\fi{\color{revtext}#1}}

\newcommand{\fix}[1]{\ifvmode\leavevmode\fi{\color{fixtext}#1}}
\newcommand{\fwmark}{\,(262k)}
\colorlet{zebra}{black!4}\colorlet{zebraov}{black!11}
\colorlet{oursrow}{black!6}
\newcommand{\tfirst}[1]{{\color{red!80!black}#1}}
\newcommand{\tsecond}[1]{{\color{blue!80!black}#1}}
\newcommand{\cmark}{\ding{51}}
\newcommand{\nomark}{{\color{black!35}--}}
\definecolor{gEnt}{HTML}{2A78D6}\definecolor{gSup}{HTML}{1BAF7A}\definecolor{gRel}{HTML}{EDA100}\definecolor{gUnr}{HTML}{9B9A94}\definecolor{gNA}{HTML}{DDDCD5}
\definecolor{vizT1}{HTML}{4A3AA7}\definecolor{vizT2}{HTML}{EB6834}\definecolor{vizPart}{HTML}{2A9D8F}
\definecolor{fDesc}{HTML}{2A78D6}\definecolor{fAsk}{HTML}{EB6834}\definecolor{fSelx}{HTML}{1BAF7A}\definecolor{fSelo}{HTML}{EDA100}\definecolor{fSelr}{HTML}{E87BA4}\colorlet{fOther}{black!30}
\definecolor{oVer}{HTML}{2A78D6}\definecolor{oKw}{HTML}{EDA100}\definecolor{oNot}{HTML}{4A3AA7}\definecolor{oUnk}{HTML}{E34948}\definecolor{oEmpty}{HTML}{DDDCD5}\definecolor{oErr}{HTML}{6E6D68}
\definecolor{provE}{HTML}{2A78D6}\definecolor{provP}{HTML}{4A3AA7}\definecolor{provT}{HTML}{1BAF7A}\definecolor{provK}{HTML}{EDA100}\definecolor{provM}{HTML}{9B9A94}\definecolor{provN}{HTML}{DDDCD5}
\definecolor{mCard}{HTML}{2A78D6}\definecolor{mEnt}{HTML}{1BAF7A}
\definecolor{qQ}{HTML}{2A78D6}\definecolor{qB}{HTML}{4A3AA7}\definecolor{qO}{HTML}{EB6834}\definecolor{qH}{HTML}{1BAF7A}\definecolor{qS}{HTML}{EDA100}\definecolor{qX}{HTML}{9B9A94}
\newcommand{\evrows}[1]{\csname @@input\endcsname #1 }
\newlength{\tOneW}\newlength{\tTwoW}

\newcommand{\loss}[1]{%
  \;{\color{red!75!black}\ensuremath{\downarrow}\,#1}%
}

\tcbset{pbox/.style={enhanced, colback=black!3, colframe=black!65, colbacktitle=black!65,
  coltitle=white, fonttitle=\small\bfseries, boxrule=.6pt, arc=2pt, left=4pt, right=4pt,
  top=2.5pt, bottom=2.5pt, toptitle=1.2pt, bottomtitle=1.2pt},
  plisting/.style={listing options={basicstyle=\ttfamily\fontsize{6.6pt}{7.7pt}\selectfont,
    breaklines=true, breakatwhitespace=false, columns=fullflexible, keepspaces=true,
    showstringspaces=false, breakindent=0pt, escapeinside={(*}{*)}}}}
\newtcolorbox{trajbox}[1]{pbox, breakable, title={#1}, before skip=6pt, after skip=6pt, fontupper=\footnotesize\raggedright, fonttitle=\footnotesize\bfseries, colupper=black}
\newtcolorbox{promptbox}[1]{pbox, breakable, title={#1}, before skip=4pt, after skip=4pt, fontupper=\footnotesize, fonttitle=\footnotesize\bfseries}
\newtcblisting{codelisting}{pbox, plisting, listing only}
\newtcbtheorem[auto counter]{exbox}{Example}{pbox, fontupper=\footnotesize, fonttitle=\footnotesize\bfseries,
  before skip=4pt, after skip=4pt, separator sign={:}, floatplacement=tbp}{ex}
\tcbset{t3case/.style={pbox, fontupper=\scriptsize\raggedright, fonttitle=\scriptsize\bfseries, colupper=black, top=1.5pt, bottom=1.5pt, toptitle=0.8pt, bottomtitle=0.8pt}}
\newcommand{\tcok}{\textcolor{green!50!black}{\ding{51}}}\newcommand{\tcno}{\textcolor{red!75!black}{\ding{55}}}
\newcommand{\tcopts}[1]{\par\vspace{1pt}{\setlength{\tabcolsep}{0pt}\begin{tabular*}{\linewidth}{@{}p{0.5\linewidth}p{0.5\linewidth}@{}}#1\end{tabular*}}\par\vspace{1pt}}
\newcommand{\tcq}[1]{\textit{``#1''}}
\newcommand{\tcomit}{\ldots}
\newcommand{\tccall}[3]{\par\noindent\textsf{#1.}~\code{#2}\par\noindent\hangindent=1.6em\hangafter=1\hspace*{0.8em}$\to$~\code{#3}\par\vspace{1pt}}
\newcommand{\slot}[1]{\textcolor{slotcolor}{#1}}
\newcommand{\qslot}[1]{\textcolor{red!75!black}{#1}}

\newcommand{\regimeA}{Formal Reasoning}
\newcommand{\regimeB}{Ontology-Grounded LLM Reasoning}
\newcommand{\regimeC}{Ontology-Assisted LLM Reasoning}
\AtEndPreamble{\theoremstyle{acmdefinition}\newtheorem*{example*}{Example}}
\begin{document}
\title{\oursname: Ontology-based Agentic RAG for Faithful LLM Reasoning}

\author{Hui Yang}
\orcid{0000-0002-4262-4001}
\affiliation{%
  \institution{The University of Manchester}
  \city{Manchester}
  \country{United Kingdom}}
\email{hui.yang-2@manchester.ac.uk}

\author{Jiaoyan Chen}
\orcid{0000-0003-4643-6750}
\affiliation{%
  \institution{The University of Manchester}
  \city{Manchester}
  \country{United Kingdom}}
\email{jiaoyan.chen@manchester.ac.uk}

\author{Yongsheng Gao}
\orcid{0000-0002-3468-2930}
\affiliation{%
  \institution{SNOMED International}
  \city{London}
  \country{United Kingdom}}
\email{yga@snomed.org}

\renewcommand{\shortauthors}{Yang et al.}

\begin{abstract}
Large language models (LLMs) perform well across many real-world tasks but remain prone to hallucination, limiting their reliability in domains such as biomedicine and healthcare, where answers must be grounded in established knowledge. Retrieval-augmented generation (RAG) mitigates this problem by incorporating external information, but often overlooks logical relationships and relies heavily on the LLM's own reasoning, which can compromise answer faithfulness. We propose \ours{}, an agentic framework that integrates formal reasoning with LLM reasoning, drawing on both the logical structure and informal annotations of domain ontologies. For each answer, \ours{} reports the ontology observations used as evidence and a grounding level (\emph{entailed}, \emph{supported}, \emph{related}, or \emph{unrelated}) that characterises the evidence's faithfulness. Across benchmarks derived from three large-scale ontologies and six real-world medical question-answering tasks, \ours{} achieves the best overall performance among the evaluated ontology-fine-tuned and popular RAG methods, with particularly strong results on reasoning-intensive tasks. The reported grounding levels also offer a useful indication of answer reliability, with stronger grounding generally associated with higher accuracy. Finally, our results demonstrate the value of ontologies as knowledge sources: \ours{} is competitive with, and sometimes outperforms, domain-specific RAG systems that draw on sources such as textbooks.
\end{abstract}

\maketitle

\section{Introduction}
\label{sec:intro}

\begin{figure}[t]
    \centering
    \def\MainIdeaWidth{0.95\linewidth}
    \input{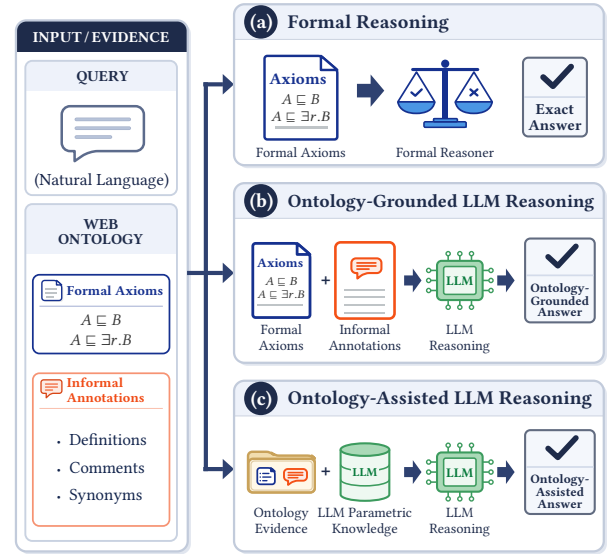}
    \caption{
    Three ontology-augmented LLM reasoning patterns, ordered from greatest to least faithfulness to the ontology:
    (a) the answer is can be inferred from the ontology’s formal axioms; (b) an LLM can infer the answer based on the ontology’s formal axioms and informal annotations; and (c) an LLM generates the answer freely, with optional grounding in relevant ontology content (classic RAG).}
    \label{fig:main-idea}
\end{figure}

Large language models (LLMs) have demonstrated strong capabilities across a wide range of knowledge-intensive tasks. However, their answers remain constrained by the knowledge encoded in their parameters and may contain hallucinations, that is, plausible but factually incorrect statements~\cite{ji2023hallucination,huang2025hallucination}.
Their reliability is further limited by the lack of transparent reasoning and by gaps in domain- or organization-specific knowledge.

Retrieval-augmented generation (RAG) augments LLMs by prompting with external evidence extracted from documents~\cite{gao2023ragsurvey}, tables~\cite{wang2024chainoftable}, and knowledge graphs~\cite{luo2024rog}, while tool-using agents enable more adaptive, dynamic and autonomous access to such sources~\cite{yao2023react,schick2023toolformer,sun2024tog,ma2025tog2}. However, existing methods mainly leverage textual and structured knowledge, leaving explicit logical knowledge under-utilised. Consequently, logical inference still relies heavily on the LLM's internal reasoning, which can make its conclusions less faithful to the given knowledge.

Meanwhile, many scientific and biomedical domains already provide rich shared, explicit and formal knowledge through ontologies in Semantic Web standards like Description Logic underpinned OWL (Web Ontology Language)~\cite{horrocks2008ontologies,smith2007obo}. These ontologies on the Web, such as the Gene Ontology (GO)~\cite{ashburner2000go}, FoodOn~\cite{dooley2018foodon}, and SNOMED~CT~\cite{donnelly2006snomed}, encode domain concepts and relations using Description Logic axioms, allowing dedicated reasoners to derive implicit relations soundly and efficiently~\cite{baader2017introduction}. Moreover, ontologies can contain rich textual annotations, including definitions and comments, providing informal but relatively important and complementary domain knowledge. This combination of formal semantics and textual annotations motivates our central question:
\begin{center}
\itshape Can ontologies bridge formal reasoning and LLM reasoning to produce more faithful answers to domain-specific questions?
\end{center}
According to the knowledge used, answering domain-specific questions utilisng LLM and ontology may include the following three reasoning patterns, as illustrated in Figure~\ref{fig:main-idea}:
\begin{itemize}[leftmargin=*]
    \item \emph{Formal reasoning}: the answer is faithful to the ontology's formal axioms and can be derived by a symbolic ontology reasoner.
    \item \emph{Ontology-grounded LLM reasoning}: the answer is faithful to the ontology's formal axioms and informal annotations (e.g., natural language labels and definitions), and requires an LLM for inference based on these ontology's information.
    \item \emph{Ontology-assisted LLM reasoning}: the answer is not faithful to the ontology, and the LLM reasoning relies on the LLM's parametric knowledge as well.
\end{itemize}
Importantly, the appropriate reasoning pattern is not known in advance and may vary across questions, requiring the system to adaptively select among the three reasoning strategies.

To address this question, we propose \ours{}, an agentic framework that combines ontology reasoning with LLM reasoning  in three stages (\S\ref{sec:method}): (1) pre-tool reasoning using the LLM’s parametric knowledge; (2) ReAct-based querying, where logical inference results and textual annotations are accessed via SPARQL; and (3) answer generation, which consolidates the evidence, and reports the ontology evidence supporting the answer with a grounding level (\S\ref{sec:grounding}).  Rather than treating the retrieved evidence as equally informative, the four levels distinguish evidence that logically entails the answer (\emph{entailed}), supports the answer without strictly entailing it (\emph{supported}), is merely relevant (\emph{related}) to the answer, or is irrelevant (\emph{unrelated}) to the answer.
This reflects the answer's faithfulness to the ontology, where entailed, supported and related correspond to formal reasoning, ontology-grounded LLM reasoning and ontology-assisted LLM reasoning in Figure \ref{fig:main-idea}, respectively.

We developed an automatic procedure for constructing evaluation datasets for the first two patterns---formal reasoning and ontology-grounded LLM reasoning---using three large-scale, real-world ontologies: GO, FoodOn, and SNOMED~CT.
For the third pattern, we evaluated \ours{} on six real-world medical question-answering tasks following the experimental setup of OntoTune~\cite{ontotune}.

Our main contributions are as follows:
\begin{itemize}[leftmargin=*,nosep]
\item We propose trustful LLM reasoning based on ontology's both formal and informal knowledge, and develop an ontology-based agentic RAG framework \ours{}, which answers a domain-specific question with evidence and their grounding levels.
\item We formulate ontology-based faithful LLM reasoning in terms of three patterns with ranging faithfulness, and construct a QA benchmark with corresponding datasets for comprehensive evaluation.
\item \ours{} achieves strong performance across all three reasoning patterns. On the first two patterns, it outperforms the strongest baseline by 10--15 percentage points on average (44.3\% vs.\ 28.9\% with Llama-3-8B-Instruct; 59.7\% vs.\ 49.7\% with Qwen3.5-9B). On the third pattern, it remains competitive with specialized medical QA methods while using $10\times$ fewer tokens than the strongest baseline with Qwen3.5-9B (9.7k vs.\ 96.7k per question).
\end{itemize}

\section{Related Work}
\label{sec:related}

\paragraph{\textbf{LLMs with external knowledge.}} Retrieval-augmented generation (RAG) augments LLMs with external knowledge from diverse sources, including unstructured text~\cite{gao2023ragsurvey} and structured sources such as tables~\cite{wang2024chainoftable} and knowledge graphs~\cite{luo2024rog}.
Domain-specific RAG has also received much attention, particularly in medicine~\cite{xiong2024medrag,xiong2024imedrag,yang2024kgrank,su2025kgarevion,wu2025medgraphrag}. More recently, growing interest has focused on RAG agents that iteratively retrieve information throughout the reasoning process~\cite{trivedi2023ircot,jiang2023flare,xiong2024imedrag}. For example, in knowledge-base question answering~\cite{lan2021survey}, graph agents traverse knowledge graphs step by step~\cite{sun2024tog,ma2025tog2}, often following the reasoning-and-acting paradigm of tool-using LLMs~\cite{yao2023react,schick2023toolformer,qin2024toollearning}.
However, these methods still rely primarily on LLMs to reason over retrieved knowledge and may therefore be less faithful and trustworthy than our approach, which additionally incorporates formal reasoning.

\paragraph{\textbf{LLMs as interfaces to formal reasoning.}} A prominent line of work combines LLMs with formal reasoning through neuro-symbolic methods, which typically translate natural language into formal representations and delegate inference to symbolic engines. Examples include executing structured queries over knowledge bases~\cite{li2023kbbinder,luo2024chatkbqa} and translating problems into logical forms for external solvers~\cite{pan2023logiclm,olausson2023linc}.
Most existing methods, however, assume a fully symbolic problem representation, where either all premises are explicitly provided or the question is entirely expressible in the vocabulary of a knowledge base. In contrast, we study a hybrid setting in which the relevant premises span formal axioms, informal annotations, and the LLM's parametric knowledge, and a question cannot necessarily be resolved by formal reasoning alone.

\paragraph{\textbf{LLMs and ontologies.}} Language models have been studied for probing factual and ontological knowledge~\cite{he2023ontolama,zhang2025ontourl}, learning ontology embeddings~\cite{he2024hit,yang2025ont}, reasoning over incomplete ontologies~\cite{yang2026neurowl}, and supporting ontology matching and learning~\cite{he2022bertmap,he2024deeponto,babaei2023llms4ol}. Retrieval-based methods have also begun to exploit ontological structure: OG-RAG~\cite{sharma2025ograg} and ORT~\cite{liu2025ort} use ontologies to structure or constrain information retrieved from documents and knowledge graphs, respectively. OntoTune~\cite{ontotune} and TaxoLLaMA~\cite{moskvoretskii2024taxollama} instead incorporate ontology knowledge into model parameters through fine-tuning. However, prior work has largely treated ontologies as sources of structured or textual knowledge, rather than as knowledge bases with both formal and informal knowledge as well as logical reasoning capabilities. To the best of our knowledge, none of the existing works explicitly integrates ontology-based logical reasoning with LLMs within an agentic framework.

\begin{figure*}
    \centering
    \def\OntologyReActWidth{0.98\linewidth}
    \input{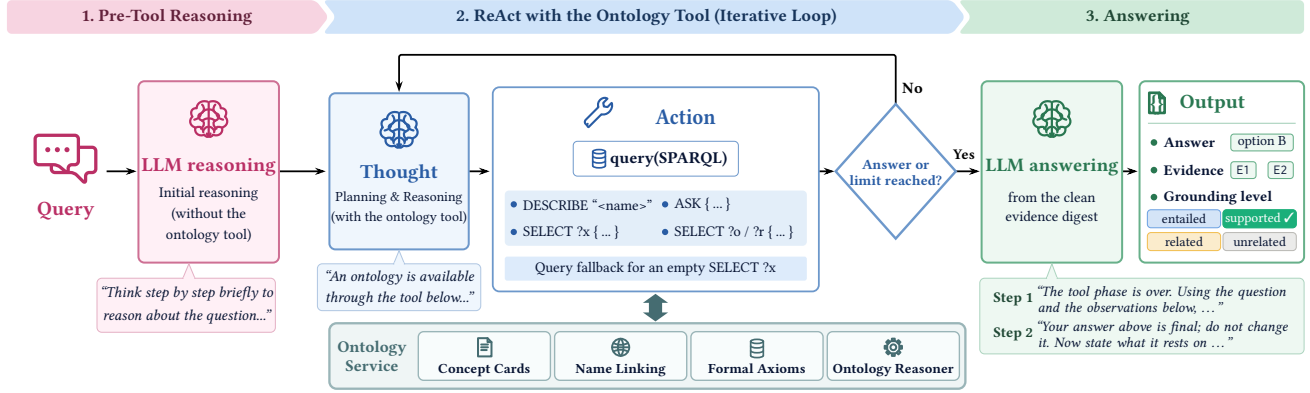}
    \caption{Overall framework of \ours: (1) pre-tool reasoning; (2) a ReAct loop of SPARQL queries to the ontology service (name linking, reasoner, query fallback); (3) answering from the evidence digest, with the cited evidence and a grounding level.}
    \label{fig:arch}
\end{figure*}

\section{Preliminaries}\label{sec:pre}

An OWL ontology~\cite{horrocks2008ontologies} represents domain knowledge using concepts (classes), relations (object properties), and logical axioms, whose semantics can be characterised by Description Logics (DLs). For the lightweight DL $\ELp$, with concept names $A\in\NC$ and role names $r\in\NR$, concepts are built by
$$C ::= \top \mid A \mid C \sqcap D \mid \exists r.C$$
and ontologies contain  General Concept Inclusions (GCIs) contain General Concept Inclusions (GCIs) $C\sqsubseteq D$, equivalences $C\equiv D$, and role inclusions $r_1\circ\cdots\circ r_n\sqsubseteq r$~\cite{DBLP:books/daglib/0041477}. We write $\Omc\models\alpha$ when the axiom $\alpha$ can be inferred from ontology $\Omc$. For example, given $\texttt{Graduate}\sqsubseteq\texttt{Student}$ and $\texttt{Student}\sqsubseteq\exists\texttt{affiliatedWith}.\texttt{Department}$, we have
$$\Omc\models\texttt{Graduate}\sqsubseteq\exists\texttt{affiliatedWith}.\texttt{Department}.$$
OWL ontologies may additionally contain \emph{annotations}, such as \texttt{rdfs:label} and \texttt{rdfs:comment}, which provide lexical and descriptive information, including alternative labels, definitions, and cross-references. Under OWL direct semantics, annotations do not affect DL entailments and are therefore ignored by DL reasoners.

Given an OWL ontology, one can construct an RDF graph consisting of $(\mathit{subject},\mathit{predicate},\mathit{object})$ triples and execute SPARQL~1.1~\cite{sparql11} queries over the graph to retrieve answers matching specified graph patterns. In particular, an axiom of the form $A\sqsubseteq\exists r.B$, whether explicitly stated in the ontology or inferred by a DL reasoner, can be materialised as a class-level triple $(A,r,B)$. Ontology annotations can likewise be represented as triples in the same graph. SPARQL can then query both types of information using graph patterns with variables: \texttt{SELECT} returns variable bindings, \texttt{ASK} tests whether a matching pattern exists, and \texttt{DESCRIBE} returns an RDF description of selected resources. These SPARQL query patterns form the basis of \ours{}.

\section{\ours{}}\label{sec:method}

We now present our main architecture, \ours, illustrated in Figure~\ref{fig:arch}. \ours consists of two main components. The first is the core agentic framework, which follows a standard ReAct~\cite{yao2023react} architecture to answer a given question through iterative reasoning and tool interaction~(\S\ref{sec:flow}). The second is the ontology service, which provides textual ontology representations and a backend that interfaces LLM-generated queries with ontology reasoners~(\S\ref{sec:service}).

\subsection{Main ReAct Architecture}
\label{sec:flow}

As illustrated in Figure~\ref{fig:arch}, \ours adopts a reasoning-and-acting architecture~\cite{yao2023react} comprising three stages: (1) pre-tool reasoning, (2) iterative reasoning and acting with access to the ontology, and (3) answer generation. The full prompts are provided in Appendix~\ref{app:prompts}.

\subsubsection{Pre-tool Reasoning}

Before entering the main ReAct loop, \ours prompts the LLM to perform an initial analysis of the input context and question. Given a pre-tool reasoning prompt $\mathcal{P}_{\mathrm{pre}}$ and a question $q$, the model $\mathcal{M}$ produces an initial reasoning state
\[
h_0=\mathcal{M}(\mathcal{P}_{\mathrm{pre}}, q).
\]

The purpose of this stage is to encourage the LLM to reason first using only the information already available in the input context and, when that information is sufficient, to form a preliminary answer. In the subsequent stage, the ontology is introduced as an additional source of structured knowledge. The model can then use the ontology to retrieve information missing from the original context and to verify, refine, or revise the preliminary answer when one has been formed.

\subsubsection{ReAct Loop}
\label{sec:loop}

Following the pre-tool reasoning stage, \ours enters the main ReAct loop, in which the model can interact with an ontology tool to retrieve ontology information and invoke the corresponding reasoner. The loop continues within the same conversation. After producing the initial reasoning state $h_0$, the model receives an agent prompt $\mathcal{P}_{\mathrm{agent}}$ as a new user turn, which introduces the available tool and specifies when it should be used.

At iteration $t$, the model generates a single action $a_t$ conditioned on the conversation history, including its previous actions and the corresponding observations returned by the ontology service  (\S\ref{sec:service}). Formally,
\begin{equation*}
a_t=\mathcal{M}\bigl(\mathcal{P}_{\mathrm{pre}},q,h_0,\mathcal{P}_{\mathrm{agent}},\{(a_j,o_j)\}_{j<t}\bigr),
\qquad
o_t=\mathcal{S}(a_t),
\end{equation*}
where $\mathcal{S}$ denotes the ontology service. The resulting observation $o_t$ is appended to the conversation and becomes part of the context for the next iteration. This process repeats until the model produces a final answer or reaches the per-item interaction budget.

\begin{promptbox}{Agent Prompt $\mathcal{P}_{\mathrm{agent}}$ (condensed)}
An ontology is available through the tool below. Continue from your answer above. You can use tools of the form \code{query(sparql)} to execute SPARQL queries over ontology concepts and relations, including \code{DESCRIBE}, \code{ASK}, and \code{SELECT} queries.

Use the ontology when the question concerns how its concepts are defined, classified, or related. For questions that do not depend on the ontology, answer using your own knowledge.
\end{promptbox}

\paragraph{Action.}
In \ours, actions are expressed as SPARQL queries~\cite{sparql11}. We support three types of queries: \texttt{DESCRIBE}, which retrieves detailed information about a concept; \texttt{ASK}, which checks whether a subsumption is entailed by the ontology reasoner; and \texttt{SELECT}, which retrieves concepts or relations satisfying specified conditions. The three action types are defined as follows:
\begin{enumerate}[leftmargin=*]
    \item \texttt{DESCRIBE "A"}:
    Given a concept name $A$, this query returns the concept card of $A$ (\S\ref{sec:docs}): its names, definition, parents, children, stated relations, and other annotations.

    \item \texttt{ASK \{"A" "r" "B"\}}:
    This query checks whether the subsumption
    $\Omc \models A \sqsubseteq \exists r.B$ holds.
    By setting $r$ to \code{"is a"} or \code{"subclass of"}, we can also check subsumptions of the form
    $\Omc \models A \sqsubseteq B$.
    The query returns \textsc{Entailed} if the entailment is verified by the ontology reasoner and \textsc{Not Entailed} otherwise. Importantly, under the open-world assumption, \textsc{Not Entailed} does not imply that the corresponding subsumption is false, since the ontology may be incomplete. We explicitly instruct the model about this distinction in the prompt.

    \item \texttt{SELECT ?x WHERE \{<conditions>\}}:
    This query retrieves concepts or relations that satisfy the specified conditions. For example, a condition of the form
    \code{\{?x "$r$" "$B$"\}}
    returns concept names $A$ such that
    $\Omc \models A \sqsubseteq \exists r.B$.
    Similarly,
    \code{\{"A" "r" ?x\}}
    asks for concepts $B$ such that
    $\Omc \models A \sqsubseteq \exists r.B$,
    while
    \code{\{"A" ?x "B"\}}
    returns relations $r$ such that
    $\Omc \models A \sqsubseteq \exists r.B$.
\end{enumerate}

\paragraph{Query fallback.}
An exact \code{SELECT ?x} query may return no results, particularly in ontology-grounded LLM reasoning, where relevant information may appear only in informal annotations, which are inaccessible to the exact reasoner. To handle such cases, we introduce a query fallback.
When a \code{SELECT ?x} query yields no matching concept, we retrieve the top-$k$ candidate concepts by comparing the query against the concept cards (\S\ref{sec:docs}) using BM25~\cite{bm25}.
\ours can then examine these candidates, inspect their associated information via \code{DESCRIBE}, and continue the ReAct loop.

\subsubsection{Answering}
Finally, we introduce a dedicated answering step that constructs a clean context from the observations gathered during reasoning. It proceeds in two steps (Figure~\ref{fig:arch}, stage~3).

\paragraph{Step 1: Answer.}
We discard observations produced by failed or uninformative tool calls, such as execution errors and queries that return empty results. If the preceding ReAct loop has already produced a candidate answer, we retain it as a draft, denoted as $\tilde y$ . Let $E$ denote the observations retained after filtering (the \emph{evidence digest}). The final answer is
\begin{equation*}
\hat{y}=\begin{cases}
\mathcal{M}\bigl(\mathcal{P}_{\mathrm{fin}},q,E,\tilde y\bigr) & \text{if } E\neq\emptyset,\\
\mathcal{M}\bigl(\mathcal{P}_{\mathrm{own}},q\bigr) & \text{otherwise.}
\end{cases}
\end{equation*}
We distinguish these two cases using separate prompts. When informative observations remain after filtering ($E\neq\emptyset$), $\mathcal{P}_{\mathrm{fin}}$ instructs the model to answer using the question, the retained observations, and the draft. Otherwise, $\mathcal{P}_{\mathrm{own}}$ asks the model to answer from its parametric knowledge.

\paragraph{Step 2: Evidence and grounding level.}\label{sec:grounding}
When the observation set $E$ is non-empty,  we continue the Step~1 conversation with one more turn, the prompt $\mathcal{P}_{\mathrm{att}}$, to identify which observations its answer relies on and to assess the strength of their support:
\begin{equation*}
(\hat C,\hat g)=\mathcal{M}\bigl(\textit{Step 1},\mathcal{P}_{\mathrm{att}}, E\bigr),
\end{equation*}
where $\hat C\subseteq E$ denotes the set of cited evidence and $\hat g$ denotes the grounding level.
We distinguish four grounding levels: (1) \emph{entailed}, when the answer can be logically derived from the facts stated in the observations; (2) \emph{supported}, when the answer can be reasonably inferred from the observations, possibly in combination with the LLM's parametric knowledge, but does not follow as a strict logical consequence; (3) \emph{related}, when the observations are relevant to the question but insufficient to determine the answer; and (4) \emph{unrelated}, when the observations are irrelevant to the question and the answer. Accordingly, we expect formal reasoning (Figure~\ref{fig:main-idea}) to produce entailed answers; ontology-grounded LLM reasoning, entailed or supported answers; and ontology-assisted LLM reasoning, mostly related answers, although unrelated, supported, and entailed cases may also occur.
Although it does not provide a formal guarantee of answer correctness, the reported grounding level reflects the model's assessment of the extent to which the available evidence supports the answer and the answer is faithful to the ontology.

\subsection{Ontology Service}
\label{sec:service}

\subsubsection{Ontology Concept Cards}%
\label{sec:docs}

To make the ontology accessible to the agent through standard text-based retrieval, we convert each concept into a \emph{concept card}, as illustrated below. Each card preserves the information in the ontology, with each item represented as a \emph{key:~value} line using the property's original label. The fields appear in a fixed order: the name, annotation definition, parents, children, stated relations, comments, and remaining annotations. All values are preserved as stated in the ontology, with three exceptions. The \emph{parents} and \emph{children} fields list the concept's inferred direct parents and children, respectively\footnote{$B$ is a direct parent of $A$ if $\Omc \models A \sqsubseteq B$ and there is no other concept $B'$ such that $\Omc \models A \sqsubseteq B' \sqsubseteq B$. Direct children are defined analogously.}. The \emph{stated relations} field lists each pair $(r,B)$ for which the ontology contains an axiom of the form $A \sqsubseteq \exists r.B \sqcap C$ or $A \equiv \exists r.B \sqcap C$, represented as ``\texttt{r -> B}''.

\begin{promptbox}{Concept card for \texttt{mitotic chromosome condensation} (GO)}
mitotic chromosome condensation

\textbf{definition:} The cell cycle process in which chromatin structure is compacted prior to and during mitosis in eukaryotic cells. \{database\_cross\_reference: GOC:mah, \ldots\}

\textbf{parents:} chromosome condensation; mitotic cell cycle process

\textbf{children:} mitotic chromosome arm condensation; \ldots

\textbf{stated relations:} part of -> mitotic cell cycle; \ldots

\textbf{database\_cross\_reference:} Reactome:R-HSA-2299718 \{rdfs:label: Condensation of Prophase Chromosomes\}; \ldots

\textbf{has\_obo\_namespace:} biological\_process

\textbf{id:} GO:0007076
\end{promptbox}

\subsubsection{Name Linking}
For a SPARQL query generated by \ours{} to be executable by the ontology service, every concept and relation name in the query must be linked to an ontology entity. We first attempt exact matching; if this fails, we retrieve candidate entities for \ours{} to inspect and use for query repair. The procedure is as follows:
\begin{itemize}[leftmargin=*]
    \item \textit{Exact match.}
    We first normalise each name (e.g., by standardising case, whitespace, and underscores) and match it against every given alternative name. If every name maps uniquely to an ontology entity, the query is executed directly. If a name matches multiple entities, it is marked as ambiguous, and the candidate entities are returned for further disambiguation. If a name cannot be matched to any ontology entity, we proceed to the fallback step.

    \item \textit{Unknown-name fallback.}
    For each concept or relation name that cannot be matched exactly, we return the five closest ontology names as candidate replacements.
    For concepts, we first retrieve a pool of 200 candidate names using BM25 and then rerank them by string similarity~\cite{ratcliff1988gestalt}. For relations, we match against the relation vocabulary directly. Given these suggestions, \ours{} can revise the query using an appropriate ontology name or reformulate the query if none of the candidates is suitable.
\end{itemize}

\section{Dataset Construction}
\label{sec:data}

We construct evaluation datasets for the two reasoning patterns in which the
ontology serves as the primary source of knowledge: \emph{formal reasoning}
(\S\ref{sec:data-formal}) and \emph{ontology-grounded LLM reasoning}
(\S\ref{sec:data-hybrid}). For ontology-assisted LLM reasoning, we do not construct a new
dataset; instead, we reuse existing domain question answering benchmarks
following~\cite{ontotune}, with details deferred to Section~\ref{sec:eval}.

\subsection{Formal Reasoning}
\label{sec:data-formal}

In the formal reasoning setting, we focus on the case where the answer to a question can be computed directly by  ontology reasoning. Specifically, we consider two kinds of questions: (1) multiple-choice questions (MCQ), where options are given and several of them may be correct; and (2) open questions (OQ), which give no options, so that the LLM has to find the answers in the ontology itself.

\subsubsection{MCQ Construction}\label{sec:data-mcq}
Our central idea is to construct questions directly from valid ontological subsumption statements by masking one component of each statement and treating the masked component as the prediction target. This process can be fully automated using an ontology reasoner. Specifically, given an entailed axiom
$$
\mathcal{O} \models A \sqsubseteq \exists r. B,
$$
we construct a query by masking either a concept, namely \(A\) or \(B\), or the relation \(r\). The masked symbol determines the reference answer, while additional concepts or relations are introduced as candidate options.

First, we obtain the entailed axioms of the form $\mathcal{O}\models A \sqsubseteq \exists r. B$ by exhaustively applying the following decomposition rule:
$$
A \sqsubseteq C\sqcap\exists r.B
\;\rightarrow\;
A \sqsubseteq \exists r. B,
\qquad
A \sqsubseteq C
$$
to the asserted axioms in \(\mathcal{O}\). Here, \(A\) and \(B\) denote atomic concepts and \(C\) denotes an arbitrary, possibly complex, \(\mathcal{EL}^+\)-concept. Then, we mask $A$, $r$, or $B$ to obtain questions. For each generated question, we construct six candidate options as follows.
\begin{itemize}[leftmargin=*]
    \item When the masked answer is a concept, we randomly select 1--3 semantically valid concepts as correct options. If \(A\) is masked, the correct options are sampled from \(A\) and its descendants; if \(B\) is masked, they are sampled from \(B\) and its ancestors. To avoid selecting concepts that are overly general or overly specific, we restrict these candidates to those within a taxonomy distance of at most two from \(A\) or \(B\), respectively. The remaining options are sampled from sibling concepts of \(A\) or \(B\), with randomly sampled concepts used when there are insufficient siblings.

    \item When the masked answer is a relation, the original relation is used as the correct option, and five distractors are sampled randomly from the other relations appearing in existential restrictions in the ontology.
\end{itemize}
Finally, to ensure the correctness of the constructed query--answer pairs, we verify each candidate option with an ontology reasoner and replace any incorrectly labeled option.

\begin{exbox}[float=t]{A generated multiple-choice item (GO, filler masked)}{mcq}
{\textbf{Source axiom.}
\emph{mesoderm formation}
\(\sqsubseteq\exists\,\)
\emph{part of}\(.\,\)
\emph{mesoderm morphogenesis}.}\\[2pt]
{\textbf{Question.}
\textit{Which concept completes this statement about ``mesoderm formation'': is part of [?]? Select the option or options that this ontology supports, and no option that is merely plausible.}}\\[2pt]
\textbf{Options.}\\
{\setlength{\tabcolsep}{0pt}\begin{tabular*}{\linewidth}{@{}p{0.56\linewidth}p{0.44\linewidth}@{}}
\textsf{A} mesoderm morphogenesis & \textsf{B} tissue morphogenesis\\
\textsf{C} morphogenesis of an epithelium & \textsf{D} cartilage morphogenesis\\
\textsf{E} fibrous ring of heart morphogenesis & \textsf{F} muscle tissue morphogenesis
\end{tabular*}}\\[2pt]
\textbf{Gold.}
{\textsf{A} and \textsf{B}: the asserted filler and its parent \emph{tissue morphogenesis}}
\end{exbox}

\subsubsection{OQ Construction}\label{sec:data-oq}
We construct an OQ \(Q\) by specifying \(n\geq 2\) conditions and asking for concepts that satisfy all of them. We restrict the search to the taxonomy branch rooted at a designated general concept \(A_0\), that is, to concepts \(A\) such that \(\Omc\models A\sqsubseteq A_0\).\footnote{For example, \(A_0\) is \emph{biological process} in GO, \emph{food product} in FoodOn, and \emph{Clinical finding} in SNOMED~CT.} An answer must additionally satisfy the relation restrictions \(\exists r_i.B_i\), for \(i=1,\ldots,n\). Thus, \(A\) is an answer to \(Q\) if
\[
\Omc\models
A\sqsubseteq
A_0
\sqcap
\exists r_1.B_1
\sqcap \cdots \sqcap
\exists r_n.B_n.
\]

We generate OQs by first selecting a concept \(A\) in the target branch and then constructing query conditions that \(A\) is known to satisfy. Specifically, for each concept \(A\) such that \(\Omc\models A\sqsubseteq A_0\), we first collect relation restrictions known to hold for \(A\), and then sample \(n\) mutually non-subsuming restrictions as the query conditions. Since all selected conditions hold for \(A\), the resulting OQ is guaranteed to have at least one answer. Appendix~\ref{app:oq} gives the details.

\begin{table*}[t]
\centering
\caption{Overall accuracy (\%). Average: mean across the three tests after within-test averaging. $^{\dagger}$: from OntoTune~\cite{ontotune}. \fwmark: full 262k-token context. \tfirst{Red} / \tsecond{blue}: best / second best.}%
\label{tab:main-compact}
\renewcommand{\arraystretch}{1.0}
\small
\setlength{\tabcolsep}{3.0pt}
\settowidth{\tOneW}{\textbf{Test 1: formal}}\setlength{\tOneW}{0.5\tOneW}\addtolength{\tOneW}{-\tabcolsep}
\settowidth{\tTwoW}{\textbf{Test 2: ontology-grounded}}\setlength{\tTwoW}{0.5\tTwoW}\addtolength{\tTwoW}{-\tabcolsep}
\newcommand{\hcell}[2]{\begin{tabular}[c]{@{}c@{}}#1\\#2\end{tabular}}%
\begin{tabular}{lcccccccccccc!{\color{white}\vrule width 1.2pt}>{\columncolor{black!8}}c}
\toprule
\multirow{2}{*}{\textbf{Method}} & \multicolumn{2}{c}{\hcell{\textbf{Knowledge}}{\textbf{Source}}} & \multicolumn{2}{c}{\hcell{\textbf{Test 1}}{(\textbf{formal})}} & \multicolumn{2}{c}{\hcell{\textbf{Test 2}}{(\textbf{ontology-grounded})}} & \multicolumn{6}{c}{\hcell{\textbf{Test 3}}{(\textbf{ontology-assisted})}} & \multicolumn{1}{c}{\multirow{2}{*}{\textbf{Average}}} \\
\cmidrule(lr){2-3}\cmidrule(lr){4-5}\cmidrule(lr){6-7}\cmidrule(lr){8-13}
 & \texttt{ont} & \texttt{doc} & MCQ & OQ & MCQ & OQ & PubMedQA & MedQA & MedMCQA & USMLE 1 & USMLE 2 & USMLE 3 & \multicolumn{1}{c}{} \\
\midrule
\multicolumn{14}{c}{\textit{Base LLMs}} \\
\cmidrule(lr){1-14}
Llama-3-8B-Instruct & \nomark & \nomark & 45.1 & 7.8 & 45.4 & 8.1 & 70.3\rlap{$^{\dagger}$} & 51.7\rlap{$^{\dagger}$} & 51.7\rlap{$^{\dagger}$} & 57.4\rlap{$^{\dagger}$} & 52.3\rlap{$^{\dagger}$} & 58.2\rlap{$^{\dagger}$} & 36.7 \\
Qwen3.5-9B & \nomark & \nomark & \tfirst{56.4} & \tfirst{8.3} & \tfirst{56.7} & \tfirst{8.6} & \tfirst{72.2} & \tfirst{78.3} & \tfirst{68.1} & \tfirst{86.2} & \tfirst{82.6} & \tfirst{89.3} & \tfirst{48.2} \\
\midrule
\multicolumn{14}{c}{\textit{Ontology Fine-Tuned LLMs (Backbone: Llama-3-8B-Instruct)}} \\
\cmidrule(lr){1-14}
OntoTune-sft & \nomark & \nomark & 44.4 & \tfirst{11.2} & 44.7 & \tfirst{11.5} & \tsecond{72.0}\rlap{$^{\dagger}$} & 51.5\rlap{$^{\dagger}$} & \tsecond{56.7}\rlap{$^{\dagger}$} & \tsecond{57.4}\rlap{$^{\dagger}$} & \tfirst{54.1}\rlap{$^{\dagger}$} & \tsecond{60.7}\rlap{$^{\dagger}$} & \tfirst{38.2} \\
OntoTune-dpo & \nomark & \nomark & \tfirst{46.4} & 6.5 & \tfirst{46.8} & 6.7 & 65.5\rlap{$^{\dagger}$} & \tfirst{53.3}\rlap{$^{\dagger}$} & \tfirst{57.2}\rlap{$^{\dagger}$} & \tfirst{58.5}\rlap{$^{\dagger}$} & \tsecond{51.4}\rlap{$^{\dagger}$} & 59.0\rlap{$^{\dagger}$} & 36.9 \\
OntoTune-sft$+$dpo & \nomark & \nomark & \tsecond{45.6} & \tsecond{9.8} & \tsecond{45.9} & \tsecond{10.1} & 66.3\rlap{$^{\dagger}$} & \tsecond{51.9}\rlap{$^{\dagger}$} & \tsecond{56.7}\rlap{$^{\dagger}$} & 53.2\rlap{$^{\dagger}$} & \tfirst{54.1}\rlap{$^{\dagger}$} & \tfirst{63.1}\rlap{$^{\dagger}$} & \tsecond{37.8} \\
TaxoLLaMA$^{*}$ & \nomark & \nomark & -- & -- & -- & -- & \tfirst{73.4}\rlap{$^{\dagger}$} & 50.5\rlap{$^{\dagger}$} & 46.1\rlap{$^{\dagger}$} & 42.6\rlap{$^{\dagger}$} & 39.4\rlap{$^{\dagger}$} & 47.5\rlap{$^{\dagger}$} & -- \\
\midrule
\multicolumn{14}{c}{\textit{Prompting and RAG Methods (Backbone: Llama-3-8B-Instruct)}} \\
\cmidrule(lr){1-14}
CoT & \nomark & \nomark & 38.2 & 3.0 & 38.5 & 3.1 & 67.7 & 55.6 & 55.3 & 64.9 & \tfirst{64.2} & \tsecond{64.8} & 34.5 \\
RAG$_{\texttt{ont}}$(text) & \cmark & \nomark & \tsecond{50.9} & 3.0 & \tfirst{53.4} & 4.6 & 65.5 & 51.7 & 51.8 & 56.4 & 56.9 & 57.4 & 37.5 \\
RAG$_{\texttt{ont}}$(query) & \cmark & \nomark & 45.3 & \tsecond{10.0} & 42.6 & 2.4 & 64.8 & 55.6 & 54.5 & 62.8 & \tsecond{57.8} & 63.9 & 36.7 \\
RAG$_{\texttt{doc}}$(text) & \nomark & \cmark & 43.0 & 2.8 & 43.3 & 2.9 & 61.6 & 54.0 & 55.5 & 56.4 & 55.0 & 55.7 & 34.1 \\
MedRAG$_{\texttt{doc}}$ & \nomark & \cmark & 40.0 & 5.7 & 40.3 & 5.8 & \tfirst{77.0} & 55.1 & 55.7 & 64.9 & 55.0 & 59.0 & 35.7 \\
ReAct$_{\texttt{ont}}$ & \cmark & \nomark & 45.0 & 9.8 & 48.1 & \tsecond{12.5} & 67.6 & 54.8 & 53.3 & 63.8 & 51.4 & 63.9 & \tsecond{38.9} \\
i-MedRAG$_{\texttt{doc}}$ & \nomark & \cmark & 38.7 & 0.5 & 38.8 & 0.5 & 66.0 & \tfirst{61.6} & \tsecond{56.2} & \tfirst{78.7} & \tfirst{64.2} & 61.5 & 34.6 \\
\rowcolor{oursrow}\textbf{\ours{}} & \cmark & \nomark & \tfirst{62.0} & \tfirst{45.0} & \tsecond{52.2} & \tfirst{18.0} & \tsecond{71.6} & \tsecond{57.6} & \tfirst{57.4} & \tsecond{66.0} & 56.9 & \tfirst{67.2} & \cellcolor{black!8}\tfirst{50.5} \\
\midrule
\multicolumn{14}{c}{\textit{Prompting and RAG Methods (Backbone: Qwen3.5-9B)}} \\
\cmidrule(lr){1-14}
CoT & \nomark & \nomark & 61.1 & 12.7 & 61.4 & 13.0 & 76.4 & 82.4 & 69.0 & 89.4 & 87.2 & \tfirst{90.2} & 52.2 \\
RAG$_{\texttt{ont}}$(text) & \cmark & \nomark & 68.7 & 18.2 & \tsecond{69.1} & 20.1 & 72.6 & 84.4 & 70.4 & \tsecond{91.5} & 82.6 & \tsecond{89.3} & 56.6 \\
RAG$_{\texttt{ont}}$(query) & \cmark & \nomark & 68.2 & 26.2 & 56.9 & 5.8 & 75.5 & 84.8 & 71.2 & \tsecond{91.5} & \tsecond{88.1} & 86.9 & 53.8 \\
RAG$_{\texttt{doc}}$(text) & \nomark & \cmark & 61.0 & 5.3 & 61.3 & 5.3 & 74.5 & 82.6 & 71.5 & \tsecond{91.5} & 81.7 & 84.4 & 49.1 \\
MedRAG$_{\texttt{doc}}$ & \nomark & \cmark & 58.9 & 0.5 & 59.2 & 0.5 & \tfirst{79.3} & 81.1 & 70.5 & \tfirst{94.7} & 79.8 & 82.8 & 47.0 \\
ReAct$_{\texttt{ont}}$ & \cmark & \nomark & \tsecond{69.2} & \tsecond{31.5} & 65.3 & \tfirst{32.9} & 75.4 & 84.7 & \tsecond{72.1} & \tsecond{91.5} & \tsecond{88.1} & 86.9 & \tsecond{60.9} \\
i-MedRAG$_{\texttt{doc}}$ & \nomark & \cmark & 40.1 & 0.5 & 40.4 & 0.5 & 71.2 & 69.8 & 66.4 & 73.4 & 65.1 & 54.1 & 35.8 \\
i-MedRAG$_{\texttt{doc}}$\fwmark & \nomark & \cmark & 40.9 & 1.0 & 41.2 & 1.0 & \tsecond{77.2} & \tfirst{88.6} & \tfirst{72.9} & \tfirst{94.7} & \tfirst{91.7} & 86.1 & 42.4 \\
\rowcolor{oursrow}\textbf{\ours{}} & \cmark & \nomark & \tfirst{79.9} & \tfirst{62.0} & \tfirst{73.7} & \tsecond{23.2} & 77.0 & \tsecond{86.3} & 71.7 & \tfirst{94.7} & \tfirst{91.7} & 87.7 & \cellcolor{black!8}\tfirst{68.1} \\
\bottomrule
\end{tabular}
\end{table*}

\begin{exbox}[float=t]{A generated open question (GO, \(n=2\))}{oq}
\textbf{Concept.} \emph{negative regulation of receptor signaling pathway via STAT}\\[2pt]
\textbf{Question.} \textit{Find one or more biological processes in this ontology that satisfy all of these conditions: negatively regulates signaling; regulates cell surface receptor signaling pathway via STAT.}\\[2pt]
\textbf{Answer set.} \emph{negative regulation of receptor signaling pathway via STAT};
\emph{negative regulation of receptor signaling pathway via JAK-STAT}; \emph{negative regulation of tyrosine phosphorylation of STAT protein}.
\end{exbox}

\subsection{Ontology-Grounded LLM Reasoning}
\label{sec:data-hybrid}

Ontology-grounded LLM reasoning targets questions that can be answered using information contained in the ontology but cannot be solved by an ontology reasoner alone. In such cases, part of the information required to answer the question is not represented as formal axioms, but instead appears in other forms, such as annotations of natural language comments and definitions. Answering these questions therefore requires combining textual evidence with the remaining logical structure of the ontology.

To construct a dataset for this setting, we reuse the same questions, options, gold answer sets, answer schemas, and sample identifiers, while replacing the original ontology with a deterministically revised version, as described below. The key idea is to move part of the information encoded in logical axioms into the concepts' natural-language definitions. As a result, the revised ontology still contains sufficient information to answer the questions when both its textual and logical content are considered, while an ontology reasoner operating only on the remaining formal axioms can no longer recover all required information.

The ontology is revised as follows. For every asserted axiom with a named concept on the left-hand side, we remove its existential restrictions from the logical layer. Specifically,
\[
A\bowtie
D\sqcap
\textstyle\bigsqcap_{i=1}^{m}\exists r_i.C_i
\text{ with }
\bowtie\in\{\sqsubseteq,\equiv\},
\quad\longmapsto\quad
A\sqsubseteq D.
\]
If no logical content remains, i.e., \(D=\top\), the revised axiom is removed entirely. Each removed restriction \(\exists r_i.C_i\) is then verbalised~\cite{he2023ontolama} and appended to the annotation definition of \(A\) (a definition is added if \(A\) has none).
We further verify that each question from Section~\ref{sec:data-formal} is answerable from the revised axioms combined with the informal annotations, but not from the axioms alone, and exclude 23 questions that fail these criteria (detailed in Appendix~\ref{app:t2checks}).

\section{{Evaluation}}
\label{sec:eval}

\begin{figure*}[t]
\centering
\definecolor{evEnt}{HTML}{2D6A4A}
\definecolor{evSup}{HTML}{5E9A7A}
\definecolor{evRel}{HTML}{D2B06A}
\definecolor{evUnr}{HTML}{DADADA}
\definecolor{evAcc}{HTML}{66717B}
\begin{tikzpicture}[x=1cm,y=1cm,font=\scriptsize,baseline=(current bounding box.north)]
  \path[use as bounding box] (0,-0.15) rectangle (17.55,-3.62);

  \node[font=\footnotesize\bfseries] at (3.65,-3.43)
    {(a) Grounding level percentage (\%)};
  \node[font=\footnotesize\bfseries] at (11.95,-3.43)
    {(b) Answer accuracy (\%)};

  \draw[black!40,line width=0.4pt,rounded corners=2pt]
    (0.25,-0.21) rectangle (5.85,-0.60);
  \foreach \x/\y/\c/\lab in {0.42/-0.405/evEnt/entailed,
                               1.74/-0.405/evSup/supported,
                               3.26/-0.405/evRel/related,
                               4.47/-0.405/evUnr/unrelated}{
    \filldraw[fill=\c,draw=black!40,line width=0.3pt]
      (\x,{\y+0.09}) rectangle ({\x+0.30},{\y-0.09});
    \node[anchor=west,inner sep=0pt,text height=1.5ex,text depth=0.25ex]
      at ({\x+0.38},\y) {\lab};
  }

  \def\cw{2.25}%
  \foreach \x/\c/\lab/\w in {7.35/evEnt/entailed/1.00,
                               9.75/evSup/supported/1.17,
                               12.15/evRel/related/0.93,
                               14.55/evUnr/unrelated/1.13}{
    \filldraw[fill=\c,draw=black!40,line width=0.3pt]
      ({\x+\cw/2-\w/2},-0.31) rectangle ({\x+\cw/2-\w/2+0.30},-0.49);
    \node[anchor=west,inner sep=0pt,text height=1.5ex,text depth=0.25ex]
      at ({\x+\cw/2-\w/2+0.38},-0.40) {\lab};
  }

  \begin{scope}[yshift=0.36cm]
  \def\sh{0.22}
  \def\sx{1.45}
  \def\sw{4.40}
  \foreach \p/\lab in {0/0,25/25,50/50,75/75,100/100}{
    \draw[black!10,line width=0.35pt]
      ({\sx+\sw*\p/100},-1.20) -- ({\sx+\sw*\p/100},-3.27);
    \node[anchor=north,inner sep=0pt,text=black!65]
      at ({\sx+\sw*\p/100},-3.33) {\lab};
  }
  \def\sharebar#1#2#3#4#5#6{
    \filldraw[fill=white,draw=black!30,line width=0.35pt]
      (\sx,#1) rectangle ({\sx+\sw},{#1-\sh});
    \fill[evEnt] (\sx,#1) rectangle ({\sx+\sw*#2/100},{#1-\sh});
    \fill[evSup] ({\sx+\sw*#2/100},#1)
      rectangle ({\sx+\sw*(#2+#3)/100},{#1-\sh});
    \fill[evRel] ({\sx+\sw*(#2+#3)/100},#1)
      rectangle ({\sx+\sw*(#2+#3+#4)/100},{#1-\sh});
    \fill[evUnr] ({\sx+\sw*(#2+#3+#4)/100},#1)
      rectangle ({\sx+\sw*(#2+#3+#4+#5)/100},{#1-\sh});
    \node[anchor=east,inner sep=0pt] at (1.32,{#1-\sh/2}) {#6};
  }

  \def\ch{0.22}
  \def\lvl#1{\ifdim#1pt<9pt\def\cc{evEnt}\def\ct{white}\def\cm{white!82!black}\else\ifdim#1pt<12pt\def\cc{evSup}\def\ct{white}\def\cm{white!82!black}%
    \else\ifdim#1pt<14pt\def\cc{evRel}\def\ct{black}\def\cm{black!60}\else\def\cc{evUnr}\def\ct{black}\def\cm{black!60}\fi\fi\fi}
  \def\cellbase#1#2#3{
    \fill[black!5] (#1,#2) rectangle ({#1+\cw},{#2-\ch});
    \draw[black!17,line width=0.25pt] (#1,#2) rectangle ({#1+\cw},{#2-\ch});
    \ifdim#3pt>0pt\filldraw[fill=\cc,draw=\cc!75!black,line width=0.25pt] (#1,#2) rectangle ({#1+\cw*#3/100},{#2-\ch});\fi
  }
  \def\accin#1#2#3#4{
    \cellbase{#1}{#2}{#3}
    \node[inner sep=0pt,text=\ct]
      at ({#1+\cw*#3/200},{#2-\ch/2}) {#4};
  }
  \def\accout#1#2#3#4{
    \cellbase{#1}{#2}{#3}
    \node[anchor=west,inner sep=0pt]
      at ({#1+\cw*#3/100+0.05},{#2-\ch/2}) {#4};
  }
  \def\accnout#1#2#3#4#5{
    \cellbase{#1}{#2}{#3}
    \node[anchor=west,inner sep=0pt]
      at ({#1+\cw*#3/100+0.05},{#2-\ch/2})
      {#4\,{\color{black!60}(m=#5)}};
  }
  \def\accnsplit#1#2#3#4#5{
    \cellbase{#1}{#2}{#3}
    \node[inner sep=0pt,text=\ct]
      at ({#1+\cw*#3/200},{#2-\ch/2}) {#4};
    \node[anchor=west,inner sep=0pt,text=black!60]
      at ({#1+\cw*#3/100+0.05},{#2-\ch/2}) {(m=#5)};
  }
  \def\accninside#1#2#3#4#5{
    \cellbase{#1}{#2}{#3}
    \node[inner sep=0pt,text=\ct]
      at ({#1+\cw*#3/200},{#2-\ch/2})
      {#4\,{\color{\cm}(m=#5)}};
  }

  \node[anchor=east,font=\scriptsize\bfseries,inner sep=0pt] at (1.32,-1.08) {Llama};
  \node[anchor=east,font=\scriptsize\bfseries,inner sep=0pt] at (7.20,-1.08) {Llama};
  \sharebar{-1.20}{83.4}{7.8}{0.9}{0.1}{Test 1}
  \sharebar{-1.52}{54.9}{27.2}{2.5}{0.1}{Test 2}
  \sharebar{-1.84}{41.9}{20.8}{0.6}{0.3}{Test 3}
  \node[anchor=east,inner sep=0pt] at (7.20,-1.31) {Test 1};
  \node[anchor=east,inner sep=0pt] at (7.20,-1.63) {Test 2};
  \node[anchor=east,inner sep=0pt] at (7.20,-1.95) {Test 3};
  \lvl{7.35}\accnsplit{7.35}{-1.20}{61.8}{61.8}{1251}
  \lvl{9.75}\accnsplit{9.75}{-1.20}{24.8}{24.8}{117}
  \lvl{12.15}\accnout{12.15}{-1.20}{7.7}{7.7}{13}
  \lvl{14.55}\accnout{14.55}{-1.20}{0.0}{0.0}{1}
  \lvl{7.35}\accnsplit{7.35}{-1.52}{45.3}{45.3}{811}
  \lvl{9.75}\accnsplit{9.75}{-1.52}{29.6}{29.6}{402}
  \lvl{12.15}\accnout{12.15}{-1.52}{5.4}{5.4}{37}
  \lvl{14.55}\accnsplit{14.55}{-1.52}{50.0}{50.0}{2}
  \lvl{7.35}\accnsplit{7.35}{-1.84}{61.3}{61.3}{2843}
  \lvl{9.75}\accnsplit{9.75}{-1.84}{58.2}{58.2}{1408}
  \lvl{12.15}\accnsplit{12.15}{-1.84}{45.5}{45.5}{44}
  \lvl{14.55}\accnsplit{14.55}{-1.84}{52.6}{52.6}{19}

  \draw[black!25,line width=0.4pt] (0.25,-2.16) -- (5.85,-2.16);
  \draw[black!25,line width=0.4pt] (6.50,-2.16) -- (16.90,-2.16);

  \node[anchor=east,font=\scriptsize\bfseries,inner sep=0pt] at (1.32,-2.30) {Qwen};
  \node[anchor=east,font=\scriptsize\bfseries,inner sep=0pt] at (7.20,-2.30) {Qwen};
  \sharebar{-2.41}{83.8}{7.7}{1.3}{0.2}{Test 1}
  \sharebar{-2.73}{62.6}{14.2}{9.0}{1.6}{Test 2}
  \sharebar{-3.05}{0.9}{38.8}{0.5}{10.9}{Test 3}
  \node[anchor=east,inner sep=0pt] at (7.20,-2.52) {Test 1};
  \node[anchor=east,inner sep=0pt] at (7.20,-2.84) {Test 2};
  \node[anchor=east,inner sep=0pt] at (7.20,-3.16) {Test 3};
  \lvl{7.35}\accninside{7.35}{-2.41}{83.0}{83.0}{1257}
  \lvl{9.75}\accnsplit{9.75}{-2.41}{34.5}{34.5}{116}
  \lvl{12.15}\accnout{12.15}{-2.41}{0.0}{0.0}{19}
  \lvl{14.55}\accnsplit{14.55}{-2.41}{33.3}{33.3}{3}
  \lvl{7.35}\accninside{7.35}{-2.73}{75.1}{75.1}{924}
  \lvl{9.75}\accnsplit{9.75}{-2.73}{40.7}{40.7}{209}
  \lvl{12.15}\accnout{12.15}{-2.73}{0.0}{0.0}{133}
  \lvl{14.55}\accnout{14.55}{-2.73}{0.0}{0.0}{23}
  \lvl{7.35}\accninside{7.35}{-3.05}{94.9}{94.9}{59}
  \lvl{9.75}\accninside{9.75}{-3.05}{82.1}{82.1}{2630}
  \lvl{12.15}\accnsplit{12.15}{-3.05}{54.3}{54.3}{35}
  \lvl{14.55}\accninside{14.55}{-3.05}{66.4}{66.4}{738}

  \draw[black!28,line width=0.5pt] (6.18,-0.53) -- (6.18,-3.27);
  \end{scope}
\end{tikzpicture}
\caption{Grounding levels reported by \ours{}. (a) Percentage of all questions at each level; note that a part of answers have no grounding levels.
(b) Answer accuracy within each grounding level; $m$ is the number of corresponding questions.}
\label{fig:ground}
\label{fig:ground-acc}
\end{figure*}
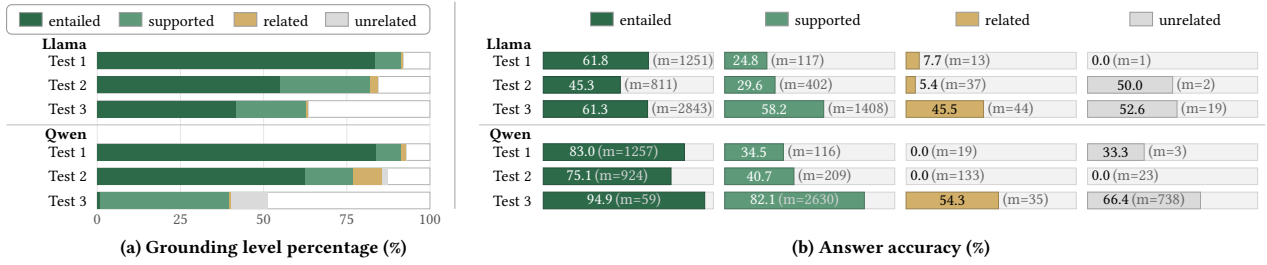

\subsection{{Experimental Setup}}
\label{sec:setup}

\textbf{Benchmarks.} We apply the procedure of Section~\ref{sec:data} to three widely used ontologies that differ in scale, terminology, and modeling conventions: GO~\cite{ashburner2000go}, FoodOn~\cite{dooley2018foodon}, and SNOMED~CT~\cite{donnelly2006snomed}, and created 300 MCQs and 200 OQs per ontology (see Appendix~\ref{app:data} for details).
\textbf{Test~1} (formal reasoning) asks these 1,500 questions against the source ontologies. \textbf{Test~2} (ontology-grounded LLM reasoning) ask 1,477 questions against the revised ontologies.
\textbf{Test~3} (ontology-assisted LLM reasoning) follows OntoTune~\cite{ontotune}: 6,781 single-answer test questions from MedQA~\cite{jin2021medqa}, USMLE Step~1/2/3~\cite{han2023medalpaca}, MedMCQA~\cite{pal2022medmcqa}, and PubMedQA~\cite{jin2019pubmedqa}, with SNOMED~CT as the underlying ontology.

\textbf{Baselines.} Besides the base LLMs, Llama-3-8B-Instruct~\cite{llama3} and Qwen3.5-9B~\cite{qwen35} (hereafter Llama and Qwen, respectively), we evaluate the following two categories of baselines:
\begin{itemize}[leftmargin=*]
    \item \emph{Ontology fine-tuned LLMs.} We include the three released OntoTune variants (sft, dpo, and sft+dpo) and TaxoLLaMA$^{*}$, OntoTune's non-public re-implementation of TaxoLLaMA~\cite{moskvoretskii2024taxollama}; all four are fine-tuned to integrate SNOMED~CT knowledge into Llama, and the Test~3 results of the Llama-based models (marked with~$\dagger$) are quoted from OntoTune.
\item \emph{Prompting and retrieval methods.}
We consider standard chain-of-thought (CoT) prompting~\cite{wei2022cot} and four retrieval-augmented methods: RAG, ReAct~\cite{yao2023react}, MedRAG~\cite{xiong2024medrag}, and i-MedRAG~\cite{xiong2024imedrag}.
We consider two retrieval sources: \textbf{(1)} \texttt{ont}, the ontology concept cards
introduced in Section~\ref{sec:docs}, and \textbf{(2)} \texttt{doc}, comprising the medical textbooks~\cite{jin2021medqa} and StatPearls~\cite{statpearls} used by MedRAG and i-MedRAG.
For RAG, we evaluate both \texttt{doc} and \texttt{ont} as retrieval sources with textual retrieval, RAG$_{\texttt{doc}}$(text) and RAG$_{\texttt{ont}}$(text), as well as a structured query variant, RAG$_{\texttt{ont}}$(query), which replaces textual retrieval with a single SPARQL query as in \ours{}.
ReAct$_{\texttt{ont}}$ uses \texttt{ont}, while MedRAG$_{\texttt{doc}}$ and i-MedRAG$_{\texttt{doc}}$ use \texttt{doc}. In particular, i-MedRAG can be viewed as a medically specialized variant of ReAct.
Implementation details are provided in Appendices~\ref{app:impl} and~\ref{app:docrag}.
\end{itemize}
For a fair comparison, we use an 8,192-token context window for all methods, matching the maximum supported by Llama. We also report i-MedRAG$_{\texttt{doc}}$ on Qwen with a 262k-token window, as its performance degrades markedly under the shorter context.

\textbf{Metrics.} We report accuracy (\%) of the all tested questions. An MCQ
question is correct if the predicted set of options exactly matches the gold set. An OQ is correct if the response is non-empty and entirely correct.

\subsection{{Main Results}}
\label{sec:main}

The overall results are shown in Table~\ref{tab:main-compact}. For methods that do not use the ontology, the Test~2 results are directly reused from Test~1. The slight performance differences are due to the 23 excluded questions described in Section~\ref{sec:data-hybrid}.

\ours{} achieves the best average accuracy with both backbones (50.5\% and 68.1\%).
On \textbf{Tests~1 (formal reasoning)} and \textbf{Tests~2 (ontology-grounded LLM reasoning)}, \ours{} outperforms all baselines in every setting except two: the Test~2 MCQs with Llama (RAG$_{\texttt{ont}}$(text), 53.4\% vs.\ 52.2\%) and the Test~2 OQs with Qwen (ReAct$_{\texttt{ont}}$, 32.9\% vs.\ 23.2\%). This is consistent with the fact that these tasks are constructed directly from ontologies and require reasoning over explicit ontological knowledge.
Notably, two of the ontology fine-tuning methods, OntoTune-sft and OntoTune-sft$+$dpo, slightly improve over their base model on the OQs (e.g., from 7.8\% to 11.2\% on Test~1 for OntoTune-sft), indicating that ontology knowledge can be partially internalised through fine-tuning. In contrast, medical RAG methods such as MedRAG$_{\texttt{doc}}$ and i-MedRAG$_{\texttt{doc}}$ fall below the base model on the OQs with both backbones, and on the MCQs too, except for MedRAG$_{\texttt{doc}}$ with Qwen. These results suggest that existing medical RAG methods may struggle when questions require explicit logical reasoning and to be answered with high faithfulness to the ontology rather than primarily retrieving relevant evidence.

On \textbf{Test~3 (ontology-assisted LLM reasoning )} , \ours{} remains competitive with the best-performing baseline, i-MedRAG$_{\texttt{doc}}$. However, with Qwen, i-MedRAG$_{\texttt{doc}}$\fwmark{} requires a substantially larger retrieval and generation budget (about $4\times$ the LLM calls and $10\times$ the tokens per question of \ours{}: 22 vs.\ 5.8 calls and 96.7k vs.\ 9.7k tokens, counting the input and generated tokens of all LLM calls of a question). Most of the difference lies in the retrieved text that the model reads, which is about $66\times$ larger for i-MedRAG$_{\texttt{doc}}$, and i-MedRAG$_{\texttt{doc}}$ also generates about $5\times$ as many tokens (Appendix~\ref{app:impl}). Moreover, when restricted to the same 8,192-token context window as \ours{}, its average Test~3 accuracy drops markedly to 66.7\%, compared with 84.9\% for \ours{}. We further find that RAG$_{\texttt{ont}}$(text) outperforms RAG$_{\texttt{doc}}$(text) on Test~3 with both backbones, and outperforms MedRAG$_{\texttt{doc}}$ with Qwen. These findings highlight the potential of ontologies as retrieval sources: compared with medical textbooks, ontologies provide more concise and structured knowledge, which may also benefit real-world QA tasks in related domains.

Beyond answer accuracy, \ours{} reports the \textbf{grounding level} of each answer (\S\ref{sec:grounding}). \textbf{Figure~\ref{fig:ground}a} shows the distribution of grounding levels in the final outputs. Because some answers lack grounding annotations (e.g., when no evidence is available), the percentages do not sum to 100\%. Entailed and supported answers dominate, accounting for 70.0\% of all questions for Llama and 53.2\% for Qwen, whereas unrelated answers are rare, comprising only 0.2\% for Llama and 7.8\% for Qwen.
\textbf{Figure~\ref{fig:ground}b} shows that, among answers labeled \emph{entailed}, \emph{supported}, or \emph{related}, stronger reported grounding is generally associated with higher accuracy. For example, with Qwen on Test~3, 94.9\% of \emph{entailed} and 82.1\% of \emph{supported} answers are correct (together accounting for 39.7\% of the questions), compared with 72.1\% of the remaining answers. The reported grounding level therefore provides a useful signal of answer reliability beyond the answer itself.
Note that  \emph{unrelated} answers are unsuitable for direct comparison, as they tend to arise from questions that are less closely aligned with the scope of the ontology. For example, some may depend primarily on reading comprehension or calculation rather than ontology-based reasoning, making their accuracy more reflective of the LLM's general capabilities.

\subsection{Analysis}
\subsubsection{Evidence Retrieval and Selection}
\label{sec:evidence}

\begin{figure}[t]
\centering
\definecolor{evC}{HTML}{2B5B8F}\definecolor{evP}{HTML}{79A4D4}\definecolor{evN}{HTML}{DADADA}\definecolor{evEnt}{HTML}{2D6A4A}\definecolor{evSup}{HTML}{5E9A7A}\definecolor{evRel}{HTML}{D2B06A}\definecolor{evUnr}{HTML}{DADADA}
\begin{tikzpicture}[x=1cm, y=1.3cm, font=\footnotesize, baseline=(current bounding box.north), ev sep/.style={white, line width=0.45pt}, ent/.style={fill=evEnt}, sup/.style={fill=evSup}, rel/.style={fill=evRel}, unr/.style={fill=evUnr}, ent changed/.style={fill=evEnt!15!white, postaction={pattern=crosshatch dots, pattern color=evEnt}}, sup changed/.style={fill=evSup!15!white, postaction={pattern=crosshatch dots, pattern color=evSup}}, rel changed/.style={fill=evRel!20!white, postaction={pattern=crosshatch dots, pattern color=evRel!80!black}}, unr changed/.style={fill=evUnr!45!white, postaction={pattern=crosshatch dots, pattern color=black!45}}, not all/.style={fill=white}]
\path (0,0.02) -- (8.40,0.02);
\node[anchor=north, draw=black!40, line width=0.4pt, rounded corners=2pt, inner xsep=4pt, inner ysep=1.5pt, font=\scriptsize] at (4.20,0.000) {%
\renewcommand{\arraystretch}{0.95}\tikzset{sharp corners}\setlength{\tabcolsep}{0pt}%
\def\chip#1{\tikz[baseline=-0.6ex]\filldraw[fill=#1, draw=black!40, line width=0.3pt] (0,-0.09) rectangle (0.30,0.09);\hspace{2.5pt}}%
\begin{tabular}{@{}l@{\hspace{6pt}}l@{}}
\textbf{Gold evidence retrieved:} & \chip{evC}completely\hspace{8pt}\chip{evP}partly\hspace{8pt}\chip{evN}not at all\\
\textbf{Retrieved evidence output:} & fully\hspace{2.5pt}\tikz[baseline=-0.6ex]{\path[fill=black!45] (0,-0.09) rectangle (0.50,0.09); \path[fill=black!8, postaction={pattern=north east lines, pattern color=black!60}] (0.50,-0.09) rectangle (1.00,0.09); \draw[black!60, line width=0.4pt] (0,-0.09) rectangle (1.00,0.09);}\hspace{2.5pt}not fully
\end{tabular}};

\node[rotate=90, font=\footnotesize\bfseries, inner sep=0pt] at (0.12,-1.090) {Llama};
\draw[black!35, line width=0.5pt] (0.30,-0.820) -- (0.30,-1.360);
\node[anchor=east, inner sep=0pt, text height=1.5ex, text depth=0.25ex] at (0.90,-0.910) {MCQ};
\path[fill=evC] (0.980,-0.820) rectangle (3.274,-1.000);
\path[fill=evC!18!white, postaction={pattern=north east lines, pattern color=evC}] (3.274,-0.820) rectangle (3.310,-1.000);
\path[fill=evP] (3.310,-0.820) rectangle (3.512,-1.000);
\path[fill=evP!18!white, postaction={pattern=north east lines, pattern color=evP}] (3.512,-0.820) rectangle (3.541,-1.000);
\path[fill=evN] (3.541,-0.820) rectangle (4.615,-1.000);
\draw[evC, line width=0.5pt] (0.980,-0.820) rectangle (3.310,-1.000);
\draw[evP, line width=0.5pt] (3.310,-0.820) rectangle (3.541,-1.000);
\draw[white, line width=0.9pt] (3.310,-0.810) -- (3.310,-1.010);
\draw[white, line width=0.9pt] (3.541,-0.810) -- (3.541,-1.010);
\node[font=\scriptsize, inner sep=0.5pt, text=white] at (2.127,-0.910) {64.1};
\node[font=\scriptsize, inner sep=0.5pt, text=black] at (4.078,-0.910) {29.6};
\path[fill=evC] (4.765,-0.820) rectangle (5.859,-1.000);
\path[fill=evC!18!white, postaction={pattern=north east lines, pattern color=evC}] (5.859,-0.820) rectangle (5.916,-1.000);
\path[fill=evP] (5.916,-0.820) rectangle (6.343,-1.000);
\path[fill=evP!18!white, postaction={pattern=north east lines, pattern color=evP}] (6.343,-0.820) rectangle (6.400,-1.000);
\path[fill=evN] (6.400,-0.820) rectangle (8.400,-1.000);
\draw[evC, line width=0.5pt] (4.765,-0.820) rectangle (5.916,-1.000);
\draw[evP, line width=0.5pt] (5.916,-0.820) rectangle (6.400,-1.000);
\draw[white, line width=0.9pt] (5.916,-0.810) -- (5.916,-1.010);
\draw[white, line width=0.9pt] (6.400,-0.810) -- (6.400,-1.010);
\node[font=\scriptsize, inner sep=0.5pt, text=white] at (5.312,-0.910) {31.7};
\node[font=\tiny, inner sep=0.5pt, text=black] at (6.129,-0.910) {13.3};
\node[font=\scriptsize, inner sep=0.5pt, text=black] at (7.400,-0.910) {55.0};

\node[anchor=east, inner sep=0pt, text height=1.5ex, text depth=0.25ex] at (0.90,-1.270) {OQ};
\path[fill=evC] (0.980,-1.180) rectangle (2.270,-1.360);
\path[fill=evC!18!white, postaction={pattern=north east lines, pattern color=evC}] (2.270,-1.180) rectangle (2.319,-1.360);
\path[fill=evP] (2.319,-1.180) rectangle (2.585,-1.360);
\path[fill=evP!18!white, postaction={pattern=north east lines, pattern color=evP}] (2.585,-1.180) rectangle (2.610,-1.360);
\path[fill=evN] (2.610,-1.180) rectangle (4.615,-1.360);
\draw[evC, line width=0.5pt] (0.980,-1.180) rectangle (2.319,-1.360);
\draw[evP, line width=0.5pt] (2.319,-1.180) rectangle (2.610,-1.360);
\draw[white, line width=0.9pt] (2.319,-1.170) -- (2.319,-1.370);
\draw[white, line width=0.9pt] (2.610,-1.170) -- (2.610,-1.370);
\node[font=\scriptsize, inner sep=0.5pt, text=white] at (1.625,-1.270) {36.8};
\node[font=\tiny, inner sep=0.5pt, text=black] at (2.464,-1.270) {8.0};
\node[font=\scriptsize, inner sep=0.5pt, text=black] at (3.612,-1.270) {55.2};
\path[fill=evP] (4.765,-1.180) rectangle (4.790,-1.360);
\path[fill=evP!18!white, postaction={pattern=north east lines, pattern color=evP}] (4.790,-1.180) rectangle (4.802,-1.360);
\path[fill=evN] (4.802,-1.180) rectangle (8.400,-1.360);
\draw[evP, line width=0.5pt] (4.765,-1.180) rectangle (4.802,-1.360);
\draw[white, line width=0.9pt] (4.802,-1.170) -- (4.802,-1.370);
\node[font=\scriptsize, inner sep=0.5pt, text=black] at (6.601,-1.270) {99.0};

\draw[black!30, line width=0.4pt] (0,-1.530) -- (8.40,-1.530);

\node[rotate=90, font=\footnotesize\bfseries, inner sep=0pt] at (0.12,-1.970) {Qwen};
\draw[black!35, line width=0.5pt] (0.30,-1.700) -- (0.30,-2.240);
\node[anchor=east, inner sep=0pt, text height=1.5ex, text depth=0.25ex] at (0.90,-1.790) {MCQ};
\path[fill=evC] (0.980,-1.700) rectangle (3.872,-1.880);
\path[fill=evC!18!white, postaction={pattern=north east lines, pattern color=evC}] (3.872,-1.700) rectangle (3.936,-1.880);
\path[fill=evP] (3.936,-1.700) rectangle (4.393,-1.880);
\path[fill=evP!18!white, postaction={pattern=north east lines, pattern color=evP}] (4.393,-1.700) rectangle (4.449,-1.880);
\path[fill=evN] (4.449,-1.700) rectangle (4.615,-1.880);
\draw[evC, line width=0.5pt] (0.980,-1.700) rectangle (3.936,-1.880);
\draw[evP, line width=0.5pt] (3.936,-1.700) rectangle (4.449,-1.880);
\draw[white, line width=0.9pt] (3.936,-1.690) -- (3.936,-1.890);
\draw[white, line width=0.9pt] (4.449,-1.690) -- (4.449,-1.890);
\node[font=\scriptsize, inner sep=0.5pt, text=white] at (2.426,-1.790) {81.3};
\node[font=\tiny, inner sep=0.5pt, text=black] at (4.165,-1.790) {14.1};
\path[fill=evC] (4.765,-1.700) rectangle (7.213,-1.880);
\path[fill=evC!18!white, postaction={pattern=north east lines, pattern color=evC}] (7.213,-1.700) rectangle (7.294,-1.880);
\path[fill=evP] (7.294,-1.700) rectangle (7.884,-1.880);
\path[fill=evP!18!white, postaction={pattern=north east lines, pattern color=evP}] (7.884,-1.700) rectangle (7.973,-1.880);
\path[fill=evN] (7.973,-1.700) rectangle (8.400,-1.880);
\draw[evC, line width=0.5pt] (4.765,-1.700) rectangle (7.294,-1.880);
\draw[evP, line width=0.5pt] (7.294,-1.700) rectangle (7.973,-1.880);
\draw[white, line width=0.9pt] (7.294,-1.690) -- (7.294,-1.890);
\draw[white, line width=0.9pt] (7.973,-1.690) -- (7.973,-1.890);
\node[font=\scriptsize, inner sep=0.5pt, text=white] at (5.989,-1.790) {69.6};
\node[font=\tiny, inner sep=0.5pt, text=black] at (7.589,-1.790) {18.7};
\node[font=\tiny, inner sep=0.5pt, text=black] at (8.187,-1.790) {11.7};

\node[anchor=east, inner sep=0pt, text height=1.5ex, text depth=0.25ex] at (0.90,-2.150) {OQ};
\path[fill=evC] (0.980,-2.060) rectangle (2.543,-2.240);
\path[fill=evC!18!white, postaction={pattern=north east lines, pattern color=evC}] (2.543,-2.060) rectangle (2.695,-2.240);
\path[fill=evP] (2.695,-2.060) rectangle (3.228,-2.240);
\path[fill=evP!18!white, postaction={pattern=north east lines, pattern color=evP}] (3.228,-2.060) rectangle (3.512,-2.240);
\path[fill=evN] (3.512,-2.060) rectangle (4.615,-2.240);
\draw[evC, line width=0.5pt] (0.980,-2.060) rectangle (2.695,-2.240);
\draw[evP, line width=0.5pt] (2.695,-2.060) rectangle (3.512,-2.240);
\draw[white, line width=0.9pt] (2.695,-2.050) -- (2.695,-2.250);
\draw[white, line width=0.9pt] (3.512,-2.050) -- (3.512,-2.250);
\node[font=\scriptsize, inner sep=0.5pt, text=white] at (1.762,-2.150) {47.2};
\node[font=\scriptsize, inner sep=0.5pt, text=black] at (2.961,-2.150) {22.5};
\node[font=\scriptsize, inner sep=0.5pt, text=black] at (4.064,-2.150) {30.3};
\path[fill=evC] (4.765,-2.060) rectangle (5.027,-2.240);
\path[fill=evC!18!white, postaction={pattern=north east lines, pattern color=evC}] (5.027,-2.060) rectangle (5.102,-2.240);
\path[fill=evP] (5.102,-2.060) rectangle (5.551,-2.240);
\path[fill=evP!18!white, postaction={pattern=north east lines, pattern color=evP}] (5.551,-2.060) rectangle (6.000,-2.240);
\path[fill=evN] (6.000,-2.060) rectangle (8.400,-2.240);
\draw[evC, line width=0.5pt] (4.765,-2.060) rectangle (5.102,-2.240);
\draw[evP, line width=0.5pt] (5.102,-2.060) rectangle (6.000,-2.240);
\draw[white, line width=0.9pt] (5.102,-2.050) -- (5.102,-2.250);
\draw[white, line width=0.9pt] (6.000,-2.050) -- (6.000,-2.250);
\node[font=\tiny, inner sep=0.5pt, text=white] at (4.933,-2.150) {9.3};
\node[font=\scriptsize, inner sep=0.5pt, text=black] at (5.326,-2.150) {24.7};
\node[font=\scriptsize, inner sep=0.5pt, text=black] at (7.200,-2.150) {66.0};

\node[anchor=north, inner sep=0pt, font=\footnotesize\bfseries] at (2.80,-2.390) {Test 1 {\tiny (formal reasoning)}};
\node[anchor=north, inner sep=0pt, font=\footnotesize\bfseries] at (6.58,-2.390) {Test 2 {\tiny (ontology-grounded LLM reasoning)}};
\end{tikzpicture}
\caption{Ratio of questions whose gold evidence is retrieved and selected (\%).}
\label{fig:ev-main-a}
\end{figure}
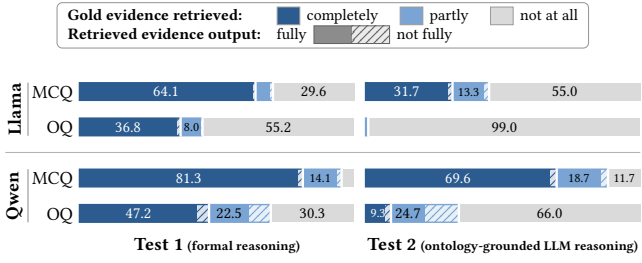

Unlike the grounding level, the \emph{gold evidence} can be traced directly from the ontology for Tests~1 and~2, allowing us to evaluate exactly whether the required evidence is retrieved and subsequently cited (Figure~\ref{fig:ev-main-a}). The results suggest that retrieval, rather than reasoning over retrieved evidence, is the primary bottleneck. MCQs achieve a substantially higher rate of fully retrieved gold evidence than OQs (61.7 vs.\ 23.6\% over all items), which is reasonable as the candidate answer options provide additional cues that constrain retrieval. Moreover, once the relevant evidence is retrieved, it is usually among the observations cited for the answer (95.1--98.4\% of the MCQs and at least 77.8\% of the OQs). Qwen also demonstrates a substantially stronger ability than Llama to retrieve useful evidence. Nevertheless, both models perform poorly on Test~2 OQs. This is expected because much of the relevant information in Test~2 is encoded in definitions and is therefore less directly accessible to the retrieval process. The absence of candidate options in OQs further increases the difficulty of identifying the relevant evidence.

\subsubsection{{Ablation Study}}
\label{sec:ablation}

To analyse the contribution of each component, we report an ablation study in Table~\ref{tab:ablation}. Across both backbones, adding the ReAct loop and the final answering step consistently improves over pre-tool reasoning alone, with the largest gains on the first two test patterns (59.7 vs.\ 36.5\% on Tests~1--2, with Qwen). This highlights the importance of ontology-grounded reasoning through SPARQL queries when the answer depends on structured ontology knowledge. The final \textbf{answering step} provides a further consistent gain (68.1 vs.\ 65.2\% on average, with Qwen), showing that synthesizing the intermediate reasoning and retrieved evidence remains important even after tool use.
The \textbf{query fallback} exhibits a more nuanced trade-off. While it improves overall performance (68.1 vs.\ 67.7\% on average, with Qwen), removing it can benefit some settings, such as Test~1 MCQs, likely because doing so avoids introducing less targeted retrieval when the structured ontology is already sufficient.

\subsubsection{{Case Study}}
\label{sec:cases}

\begin{table}[t]
\centering
\caption{Ablation of \ours{} (accuracy \%): each indented row changes one component;
\loss: drop of percentage points over the original OA-RAG.}
\label{tab:ablation}
\renewcommand{\arraystretch}{1.0}
\small
\setlength{\tabcolsep}{2.4pt}
\begin{tabular}{@{}lcc@{\hspace{5pt}}cc@{\hspace{5pt}}c@{\hspace{4pt}}>{\columncolor{black!8}[\tabcolsep][0pt]}c@{}>{\columncolor{black!8}[0pt][\tabcolsep]}l@{}}
\toprule
\multirow{2}{*}{\textbf{Variant}} & \multicolumn{2}{c}{\textbf{Test 1}} & \multicolumn{2}{c}{\textbf{Test 2}} & \multirow{2}{*}{\textbf{Test 3}} & \multicolumn{2}{c}{\multirow{2}{*}{\textbf{Average}}} \\
\cmidrule(lr){2-3}\cmidrule(lr){4-5}
 & MCQ & OQ & MCQ & OQ &  & \multicolumn{2}{c}{} \\
\midrule
\multicolumn{8}{c}{\textit{Backbone: Llama-3-8B-Instruct}} \\
\cmidrule(lr){1-8}
\ours{} & \tsecond{62.0} & \tfirst{45.0} & \tsecond{52.2} & \tsecond{18.0} & \tfirst{62.8} & \tfirst{50.5} &  \\
\quad pre-tool reasoning only & 43.4 & 4.8 & 43.8 & 4.5 & \tsecond{61.9} & 36.7 & \loss{13.8} \\
\quad w/o answering step & 57.3 & 38.7 & 48.0 & 7.9 & 58.1 & 44.7 & \loss{5.8} \\
\quad w/o query fallback & \tfirst{62.1} & \tsecond{42.7} & \tfirst{52.6} & 9.3 & \tfirst{62.8} & \tsecond{48.7} & \loss{1.8} \\
\midrule
\multicolumn{8}{c}{\textit{Backbone: Qwen3.5-9B}} \\
\cmidrule(lr){1-8}
\ours{} & 79.9 & \tfirst{62.0} & \tsecond{73.7} & \tsecond{23.2} & \tsecond{84.9} & \tfirst{68.1} &  \\
\quad pre-tool reasoning only & 63.6 & 9.2 & 63.9 & 9.4 & 84.7 & 52.6 & \loss{15.5} \\
\quad w/o answering step & 78.0 & 54.0 & 73.3 & 16.5 & 84.7 & 65.2 & \loss{2.9} \\
\quad w/o query fallback & \tsecond{80.4} & \tsecond{61.7} & \tfirst{75.8} & 17.8 & \tfirst{85.1} & \tsecond{67.7} & \loss{0.4} \\
\bottomrule
\end{tabular}
\end{table}

\begin{figure}[t]
\centering
\begin{tcolorbox}[t3case, title={(a) The ontology names halothane as the causative agent \hfill MedQA}]
\textbf{Q:} Five days after undergoing an emergency appendectomy under general inhalational anesthesia \tcomit\ she received a transfusion of 1 unit of packed red blood cells \tcomit\ Examination shows jaundice \tcomit\ Eosinophils 13\% \tcomit\ A biopsy of the liver shows massive centrilobular necrosis. Which of the following is the most likely underlying cause \tcomit?
\tcopts{\textbf{(A) Adverse effect of anesthetic} & (B) Gram negative bacteria in the bloodstream\\ (C) Trauma to the bile duct & (D) Acalculous inflammation of the gallbladder}
(E) Excessive lysis of red blood cells\par\vspace{1pt}
\textbf{Pre-tool:} \tcq{\tcomit\ The hemolysis releases free hemoglobin, which is toxic to the liver \tcomit} $\Rightarrow$ E \tcno
\par\vspace{1pt}
\textbf{Tool phase} (call 4 of 4):
\tccall{4}{DESCRIBE "Halothane hepatitis"}{Halothane hepatitis \tcomit\ parents: Drug-induced hepatitis; Poisoning caused by halothane \tcomit\ stated relations: \tcomit; Causative agent -> Halothane; Finding site -> Liver structure}
\textbf{Answer:} \tcq{\tcomit\ is diagnostic of Halothane hepatitis, which is an adverse effect of the anesthetic} $\Rightarrow$ A \tcok
\end{tcolorbox}
\vspace{3pt}
\begin{tcolorbox}[t3case, title={(b) The ontology maps blind staggers to selenium \hfill MedMCQA}]
\textbf{Q:} Poisoning by which of these elements causes blind staggers alkali disease in livestock?
\tcopts{(A) Magnesium & (B) Manganese\\ \textbf{(C) Selenium} & (D) Molybdenum}
\textbf{Pre-tool:} \tcq{\tcomit\ Molybdenum: Toxicity (or high levels relative to sulfur/copper) is the direct cause of the condition described as alkali disease leading to blind staggers} $\Rightarrow$ D \tcno
\par\vspace{1pt}
\textbf{Tool phase} (calls 1 and 3 of 3):
\tccall{1}{DESCRIBE "blind staggers"}{Selenium poisoning \tcomit\ stated relations: Causative agent -> Selenium; skos:altLabel: Alkali disease; Blind staggers}
\tccall{3}{ASK \{ "blind staggers" "Causative agent" "Selenium" \}}{every Selenium poisoning is related by Causative agent to some Selenium: ENTAILED}
\textbf{Answer:} \tcq{\tcomit\ The provided observations explicitly link `Selenium poisoning' with the alternative labels `Alkali disease' and `Blind staggers' and identify Selenium as the causative agent} $\Rightarrow$ C \tcok
\end{tcolorbox}
\caption{Two Test~3 trajectories of \ours{} (Qwen), with intermediate reasoning omitted for brevity.}
\label{fig:t3cases}
\end{figure}

Figure~\ref{fig:t3cases} illustrates a common correction pattern: the model's pre-tool answer is plausible but based on a loose association, whereas the ontology provides an explicit relation that directly supports the correct option. In case (a), the model attributes the liver injury to hemolysis after the transfusion, whereas the ontology describes \emph{Halothane hepatitis} as a drug-induced hepatitis whose causative agent is halothane, leading to A. In case (b), the model associates \emph{blind staggers} with molybdenum, while the ontology identifies it as an alternative name for \emph{Selenium poisoning} and directly links the condition to selenium, leading to C. These cases show that the tool phase is most useful when the ontology contains a direct, question-relevant relation that can override an incorrect model association.

\section{Conclusion}
\label{sec:conclusion}

We presented \ours{}, which, to our knowledge, is the first agentic RAG framework that combines LLM reasoning with formal OWL ontology reasoning. It can adaptively address three distinct reasoning patterns: formal reasoning, ontology-grounded LLM reasoning, and ontology-assisted LLM reasoning. Experiments across all three patterns with QA over three large-scale real-world ontologies demonstrate its strong and consistent performance. They also show that ontologies provide an effective and efficient knowledge resource, with competitive performance but substantially fewer input and generated tokens in comparison with textual resources.

For future work, we plan to extend \ours{} toward agentic web that coordinates knowledge across multiple ontologies, supporting more complex questions that require reasoning across domains.

\bibliographystyle{ACM-Reference-Format}
\bibliography{sample-base}

\begin{thebibliography}{61}

\ifx \showCODEN    \undefined \def \showCODEN     #1{\unskip}     \fi
\ifx \showISBNx    \undefined \def \showISBNx     #1{\unskip}     \fi
\ifx \showISBNxiii \undefined \def \showISBNxiii  #1{\unskip}     \fi
\ifx \showISSN     \undefined \def \showISSN      #1{\unskip}     \fi
\ifx \showLCCN     \undefined \def \showLCCN      #1{\unskip}     \fi
\ifx \shownote     \undefined \def \shownote      #1{#1}          \fi
\ifx \showarticletitle \undefined \def \showarticletitle #1{#1}   \fi
\ifx \showURL      \undefined \def \showURL       {\relax}        \fi
\providecommand\bibfield[2]{#2}
\providecommand\bibinfo[2]{#2}
\providecommand\natexlab[1]{#1}
\providecommand\showeprint[2][]{arXiv:#2}

\bibitem[Ashburner et~al\mbox{.}(2000)]%
        {ashburner2000go}
\bibfield{author}{\bibinfo{person}{Michael Ashburner}, \bibinfo{person}{Catherine~A. Ball}, \bibinfo{person}{Judith~A. Blake}, \bibinfo{person}{David Botstein}, \bibinfo{person}{Heather Butler}, \bibinfo{person}{J.~Michael Cherry}, {et~al\mbox{.}}} \bibinfo{year}{2000}\natexlab{}.
\newblock \showarticletitle{{Gene Ontology: Tool for the Unification of Biology}}.
\newblock \bibinfo{journal}{\emph{Nature Genetics}} \bibinfo{volume}{25}, \bibinfo{number}{1} (\bibinfo{year}{2000}), \bibinfo{pages}{25--29}.
\newblock
\href{https://doi.org/10.1038/75556}{doi:\nolinkurl{10.1038/75556}}

\bibitem[Baader et~al\mbox{.}(2017a)]%
        {DBLP:books/daglib/0041477}
\bibfield{author}{\bibinfo{person}{Franz Baader}, \bibinfo{person}{Ian Horrocks}, \bibinfo{person}{Carsten Lutz}, {and} \bibinfo{person}{Ulrike Sattler}.} \bibinfo{year}{2017}\natexlab{a}.
\newblock \bibinfo{booktitle}{\emph{An Introduction to Description Logic}}.
\newblock \bibinfo{publisher}{Cambridge University Press}.
\newblock
\showISBNx{978-0-521-69542-8}

\bibitem[Baader et~al\mbox{.}(2017b)]%
        {baader2017introduction}
\bibfield{author}{\bibinfo{person}{Franz Baader}, \bibinfo{person}{Ian Horrocks}, \bibinfo{person}{Carsten Lutz}, \bibinfo{person}{Uli Sattler}, {et~al\mbox{.}}} \bibinfo{year}{2017}\natexlab{b}.
\newblock \bibinfo{booktitle}{\emph{An introduction to description logic}}. Vol.~\bibinfo{volume}{24}.
\newblock \bibinfo{publisher}{Cambridge University Press Cambridge}.
\newblock

\bibitem[Babaei~Giglou et~al\mbox{.}(2023)]%
        {babaei2023llms4ol}
\bibfield{author}{\bibinfo{person}{Hamed Babaei~Giglou}, \bibinfo{person}{Jennifer D'Souza}, {and} \bibinfo{person}{S{\"o}ren Auer}.} \bibinfo{year}{2023}\natexlab{}.
\newblock \showarticletitle{{LLMs4OL: Large Language Models for Ontology Learning}}. In \bibinfo{booktitle}{\emph{ISWC}}. \bibinfo{pages}{408--427}.
\newblock

\bibitem[Balhoff and Mungall(2024)]%
        {balhoff2024whelk}
\bibfield{author}{\bibinfo{person}{James~P. Balhoff} {and} \bibinfo{person}{Christopher~J. Mungall}.} \bibinfo{year}{2024}\natexlab{}.
\newblock \showarticletitle{{Whelk: An OWL EL+RL Reasoner Enabling New Use Cases}}.
\newblock \bibinfo{journal}{\emph{Transactions on Graph Data and Knowledge}} \bibinfo{volume}{2}, \bibinfo{number}{2} (\bibinfo{year}{2024}), \bibinfo{pages}{7:1--7:17}.
\newblock
\href{https://doi.org/10.4230/TGDK.2.2.7}{doi:\nolinkurl{10.4230/TGDK.2.2.7}}

\bibitem[Cormack et~al\mbox{.}(2009)]%
        {rrf}
\bibfield{author}{\bibinfo{person}{Gordon~V. Cormack}, \bibinfo{person}{Charles L.~A. Clarke}, {and} \bibinfo{person}{Stefan Buettcher}.} \bibinfo{year}{2009}\natexlab{}.
\newblock \showarticletitle{{Reciprocal Rank Fusion Outperforms Condorcet and Individual Rank Learning Methods}}. In \bibinfo{booktitle}{\emph{SIGIR}}. \bibinfo{pages}{758--759}.
\newblock
\href{https://doi.org/10.1145/1571941.1572114}{doi:\nolinkurl{10.1145/1571941.1572114}}

\bibitem[Donnelly(2006)]%
        {donnelly2006snomed}
\bibfield{author}{\bibinfo{person}{Kevin Donnelly}.} \bibinfo{year}{2006}\natexlab{}.
\newblock \showarticletitle{{SNOMED-CT: The Advanced Terminology and Coding System for eHealth}}.
\newblock \bibinfo{journal}{\emph{Studies in Health Technology and Informatics}}  \bibinfo{volume}{121} (\bibinfo{year}{2006}), \bibinfo{pages}{279--290}.
\newblock

\bibitem[Dooley et~al\mbox{.}(2018)]%
        {dooley2018foodon}
\bibfield{author}{\bibinfo{person}{Damion~M. Dooley}, \bibinfo{person}{Emma~J. Griffiths}, \bibinfo{person}{Gurinder~S. Gosal}, \bibinfo{person}{Pier~L. Buttigieg}, \bibinfo{person}{Robert Hoehndorf}, \bibinfo{person}{Matthew~C. Lange}, \bibinfo{person}{Lynn~M. Schriml}, \bibinfo{person}{Fiona S.~L. Brinkman}, {and} \bibinfo{person}{William W.~L. Hsiao}.} \bibinfo{year}{2018}\natexlab{}.
\newblock \showarticletitle{{FoodOn: A Harmonized Food Ontology to Increase Global Food Traceability, Quality Control and Data Integration}}.
\newblock \bibinfo{journal}{\emph{npj Science of Food}}  \bibinfo{volume}{2} (\bibinfo{year}{2018}), \bibinfo{pages}{23}.
\newblock
\href{https://doi.org/10.1038/s41538-018-0032-6}{doi:\nolinkurl{10.1038/s41538-018-0032-6}}

\bibitem[Gao et~al\mbox{.}(2023)]%
        {gao2023ragsurvey}
\bibfield{author}{\bibinfo{person}{Yunfan Gao}, \bibinfo{person}{Yun Xiong}, \bibinfo{person}{Xinyu Gao}, \bibinfo{person}{Kangxiang Jia}, \bibinfo{person}{Jinliu Pan}, \bibinfo{person}{Yuxi Bi}, \bibinfo{person}{Yi Dai}, \bibinfo{person}{Jiawei Sun}, \bibinfo{person}{Meng Wang}, {and} \bibinfo{person}{Haofen Wang}.} \bibinfo{year}{2023}\natexlab{}.
\newblock \bibinfo{title}{{Retrieval-Augmented Generation for Large Language Models: A Survey}}.
\newblock
\newblock
\shownote{arXiv:2312.10997}.

\bibitem[Glimm et~al\mbox{.}(2014)]%
        {glimm2014hermit}
\bibfield{author}{\bibinfo{person}{Birte Glimm}, \bibinfo{person}{Ian Horrocks}, \bibinfo{person}{Boris Motik}, \bibinfo{person}{Giorgos Stoilos}, {and} \bibinfo{person}{Zhe Wang}.} \bibinfo{year}{2014}\natexlab{}.
\newblock \showarticletitle{{HermiT: An OWL 2 Reasoner}}.
\newblock \bibinfo{journal}{\emph{Journal of Automated Reasoning}} \bibinfo{volume}{53}, \bibinfo{number}{3} (\bibinfo{year}{2014}), \bibinfo{pages}{245--269}.
\newblock
\href{https://doi.org/10.1007/s10817-014-9305-1}{doi:\nolinkurl{10.1007/s10817-014-9305-1}}

\bibitem[Grattafiori et~al\mbox{.}(2024)]%
        {llama3}
\bibfield{author}{\bibinfo{person}{Aaron Grattafiori}, \bibinfo{person}{Abhimanyu Dubey}, {et~al\mbox{.}}} \bibinfo{year}{2024}\natexlab{}.
\newblock \bibinfo{title}{{The Llama 3 Herd of Models}}.
\newblock
\newblock
\shownote{arXiv:2407.21783}.

\bibitem[Han et~al\mbox{.}(2023)]%
        {han2023medalpaca}
\bibfield{author}{\bibinfo{person}{Tianyu Han}, \bibinfo{person}{Lisa~C. Adams}, \bibinfo{person}{Jens-Michalis Papaioannou}, \bibinfo{person}{Paul Grundmann}, \bibinfo{person}{Tom Oberhauser}, \bibinfo{person}{Alexei Figueroa}, \bibinfo{person}{Alexander L{\"o}ser}, \bibinfo{person}{Daniel Truhn}, {and} \bibinfo{person}{Keno~K. Bressem}.} \bibinfo{year}{2023}\natexlab{}.
\newblock \bibinfo{title}{{MedAlpaca -- An Open-Source Collection of Medical Conversational AI Models and Training Data}}.
\newblock
\newblock
\shownote{arXiv:2304.08247}.

\bibitem[Harris and Seaborne(2013)]%
        {sparql11}
\bibfield{author}{\bibinfo{person}{Steve Harris} {and} \bibinfo{person}{Andy Seaborne}.} \bibinfo{year}{2013}\natexlab{}.
\newblock \bibinfo{title}{{SPARQL 1.1 Query Language}}.
\newblock
\newblock
\shownote{W3C Recommendation, 21 March 2013. \url{https://www.w3.org/TR/sparql11-query/}}.

\bibitem[He et~al\mbox{.}(2022)]%
        {he2022bertmap}
\bibfield{author}{\bibinfo{person}{Yuan He}, \bibinfo{person}{Jiaoyan Chen}, \bibinfo{person}{Denvar Antonyrajah}, {and} \bibinfo{person}{Ian Horrocks}.} \bibinfo{year}{2022}\natexlab{}.
\newblock \showarticletitle{{BERTMap: A BERT-Based Ontology Alignment System}}. In \bibinfo{booktitle}{\emph{AAAI}}. \bibinfo{pages}{5684--5691}.
\newblock

\bibitem[He et~al\mbox{.}(2024a)]%
        {he2024deeponto}
\bibfield{author}{\bibinfo{person}{Yuan He}, \bibinfo{person}{Jiaoyan Chen}, \bibinfo{person}{Hang Dong}, \bibinfo{person}{Ian Horrocks}, \bibinfo{person}{Carlo Allocca}, \bibinfo{person}{Taehun Kim}, {and} \bibinfo{person}{Brahmananda Sapkota}.} \bibinfo{year}{2024}\natexlab{a}.
\newblock \showarticletitle{{DeepOnto: A Python Package for Ontology Engineering with Deep Learning}}.
\newblock \bibinfo{journal}{\emph{Semantic Web}} \bibinfo{volume}{15}, \bibinfo{number}{5} (\bibinfo{year}{2024}), \bibinfo{pages}{1991--2004}.
\newblock

\bibitem[He et~al\mbox{.}(2023)]%
        {he2023ontolama}
\bibfield{author}{\bibinfo{person}{Yuan He}, \bibinfo{person}{Jiaoyan Chen}, \bibinfo{person}{Ernesto Jim{\'e}nez-Ruiz}, \bibinfo{person}{Hang Dong}, {and} \bibinfo{person}{Ian Horrocks}.} \bibinfo{year}{2023}\natexlab{}.
\newblock \showarticletitle{{Language Model Analysis for Ontology Subsumption Inference}}. In \bibinfo{booktitle}{\emph{Findings of ACL}}. \bibinfo{pages}{3439--3453}.
\newblock
\href{https://doi.org/10.18653/v1/2023.findings-acl.213}{doi:\nolinkurl{10.18653/v1/2023.findings-acl.213}}

\bibitem[He et~al\mbox{.}(2024b)]%
        {he2024hit}
\bibfield{author}{\bibinfo{person}{Yuan He}, \bibinfo{person}{Zhangdie Yuan}, \bibinfo{person}{Jiaoyan Chen}, {and} \bibinfo{person}{Ian Horrocks}.} \bibinfo{year}{2024}\natexlab{b}.
\newblock \showarticletitle{{Language Models as Hierarchy Encoders}}. In \bibinfo{booktitle}{\emph{NeurIPS}}, Vol.~\bibinfo{volume}{37}. \bibinfo{pages}{14690--14711}.
\newblock

\bibitem[Horrocks(2008)]%
        {horrocks2008ontologies}
\bibfield{author}{\bibinfo{person}{Ian Horrocks}.} \bibinfo{year}{2008}\natexlab{}.
\newblock \showarticletitle{{Ontologies and the Semantic Web}}.
\newblock \bibinfo{journal}{\emph{Commun. ACM}} \bibinfo{volume}{51}, \bibinfo{number}{12} (\bibinfo{year}{2008}), \bibinfo{pages}{58--67}.
\newblock
\href{https://doi.org/10.1145/1409360.1409377}{doi:\nolinkurl{10.1145/1409360.1409377}}

\bibitem[Huang et~al\mbox{.}(2025)]%
        {huang2025hallucination}
\bibfield{author}{\bibinfo{person}{Lei Huang}, \bibinfo{person}{Weijiang Yu}, \bibinfo{person}{Weitao Ma}, \bibinfo{person}{Weihong Zhong}, \bibinfo{person}{Zhangyin Feng}, \bibinfo{person}{Haotian Wang}, \bibinfo{person}{Qianglong Chen}, \bibinfo{person}{Weihua Peng}, \bibinfo{person}{Xiaocheng Feng}, \bibinfo{person}{Bing Qin}, {and} \bibinfo{person}{Ting Liu}.} \bibinfo{year}{2025}\natexlab{}.
\newblock \showarticletitle{{A Survey on Hallucination in Large Language Models: Principles, Taxonomy, Challenges, and Open Questions}}.
\newblock \bibinfo{journal}{\emph{ACM Transactions on Information Systems}} \bibinfo{volume}{43}, \bibinfo{number}{2} (\bibinfo{year}{2025}), \bibinfo{pages}{1--55}.
\newblock
\href{https://doi.org/10.1145/3703155}{doi:\nolinkurl{10.1145/3703155}}

\bibitem[Ji et~al\mbox{.}(2023)]%
        {ji2023hallucination}
\bibfield{author}{\bibinfo{person}{Ziwei Ji}, \bibinfo{person}{Nayeon Lee}, \bibinfo{person}{Rita Frieske}, \bibinfo{person}{Tiezheng Yu}, \bibinfo{person}{Dan Su}, \bibinfo{person}{Yan Xu}, \bibinfo{person}{Etsuko Ishii}, \bibinfo{person}{Ye~Jin Bang}, \bibinfo{person}{Andrea Madotto}, {and} \bibinfo{person}{Pascale Fung}.} \bibinfo{year}{2023}\natexlab{}.
\newblock \showarticletitle{{Survey of Hallucination in Natural Language Generation}}.
\newblock \bibinfo{journal}{\emph{Comput. Surveys}} \bibinfo{volume}{55}, \bibinfo{number}{12} (\bibinfo{year}{2023}), \bibinfo{pages}{1--38}.
\newblock

\bibitem[Jiang et~al\mbox{.}(2023)]%
        {jiang2023flare}
\bibfield{author}{\bibinfo{person}{Zhengbao Jiang}, \bibinfo{person}{Frank~F. Xu}, \bibinfo{person}{Luyu Gao}, \bibinfo{person}{Zhiqing Sun}, \bibinfo{person}{Qian Liu}, \bibinfo{person}{Jane Dwivedi-Yu}, \bibinfo{person}{Yiming Yang}, \bibinfo{person}{Jamie Callan}, {and} \bibinfo{person}{Graham Neubig}.} \bibinfo{year}{2023}\natexlab{}.
\newblock \showarticletitle{{Active Retrieval Augmented Generation}}. In \bibinfo{booktitle}{\emph{EMNLP}}. \bibinfo{pages}{7969--7992}.
\newblock

\bibitem[Jin et~al\mbox{.}(2021)]%
        {jin2021medqa}
\bibfield{author}{\bibinfo{person}{Di Jin}, \bibinfo{person}{Eileen Pan}, \bibinfo{person}{Nassim Oufattole}, \bibinfo{person}{Wei-Hung Weng}, \bibinfo{person}{Hanyi Fang}, {and} \bibinfo{person}{Peter Szolovits}.} \bibinfo{year}{2021}\natexlab{}.
\newblock \showarticletitle{{What Disease Does This Patient Have? A Large-Scale Open Domain Question Answering Dataset from Medical Exams}}.
\newblock \bibinfo{journal}{\emph{Applied Sciences}} \bibinfo{volume}{11}, \bibinfo{number}{14} (\bibinfo{year}{2021}), \bibinfo{pages}{6421}.
\newblock
\href{https://doi.org/10.3390/app11146421}{doi:\nolinkurl{10.3390/app11146421}}

\bibitem[Jin et~al\mbox{.}(2019)]%
        {jin2019pubmedqa}
\bibfield{author}{\bibinfo{person}{Qiao Jin}, \bibinfo{person}{Bhuwan Dhingra}, \bibinfo{person}{Zhengping Liu}, \bibinfo{person}{William Cohen}, {and} \bibinfo{person}{Xinghua Lu}.} \bibinfo{year}{2019}\natexlab{}.
\newblock \showarticletitle{{PubMedQA: A Dataset for Biomedical Research Question Answering}}. In \bibinfo{booktitle}{\emph{EMNLP-IJCNLP}}. \bibinfo{pages}{2567--2577}.
\newblock
\href{https://doi.org/10.18653/v1/D19-1259}{doi:\nolinkurl{10.18653/v1/D19-1259}}

\bibitem[Jin et~al\mbox{.}(2023)]%
        {jin2023medcpt}
\bibfield{author}{\bibinfo{person}{Qiao Jin}, \bibinfo{person}{Won Kim}, \bibinfo{person}{Qingyu Chen}, \bibinfo{person}{Donald~C. Comeau}, \bibinfo{person}{Lana Yeganova}, \bibinfo{person}{W.~John Wilbur}, {and} \bibinfo{person}{Zhiyong Lu}.} \bibinfo{year}{2023}\natexlab{}.
\newblock \showarticletitle{{MedCPT: Contrastive Pre-trained Transformers with Large-scale PubMed Search Logs for Zero-shot Biomedical Information Retrieval}}.
\newblock \bibinfo{journal}{\emph{Bioinformatics}} \bibinfo{volume}{39}, \bibinfo{number}{11} (\bibinfo{year}{2023}), \bibinfo{pages}{btad651}.
\newblock

\bibitem[Kazakov et~al\mbox{.}(2014)]%
        {kazakov2014elk}
\bibfield{author}{\bibinfo{person}{Yevgeny Kazakov}, \bibinfo{person}{Markus Kr{\"o}tzsch}, {and} \bibinfo{person}{Franti{\v{s}}ek Siman{\v{c}}{\'\i}k}.} \bibinfo{year}{2014}\natexlab{}.
\newblock \showarticletitle{{The Incredible ELK: From Polynomial Procedures to Efficient Reasoning with EL Ontologies}}.
\newblock \bibinfo{journal}{\emph{Journal of Automated Reasoning}} \bibinfo{volume}{53}, \bibinfo{number}{1} (\bibinfo{year}{2014}), \bibinfo{pages}{1--61}.
\newblock
\href{https://doi.org/10.1007/s10817-013-9296-3}{doi:\nolinkurl{10.1007/s10817-013-9296-3}}

\bibitem[Lan et~al\mbox{.}(2021)]%
        {lan2021survey}
\bibfield{author}{\bibinfo{person}{Yunshi Lan}, \bibinfo{person}{Gaole He}, \bibinfo{person}{Jinhao Jiang}, \bibinfo{person}{Jing Jiang}, \bibinfo{person}{Wayne~Xin Zhao}, {and} \bibinfo{person}{Ji-Rong Wen}.} \bibinfo{year}{2021}\natexlab{}.
\newblock \showarticletitle{{A Survey on Complex Knowledge Base Question Answering: Methods, Challenges and Solutions}}. In \bibinfo{booktitle}{\emph{IJCAI}}. \bibinfo{pages}{4483--4491}.
\newblock
\href{https://doi.org/10.24963/ijcai.2021/611}{doi:\nolinkurl{10.24963/ijcai.2021/611}}

\bibitem[Li et~al\mbox{.}(2023)]%
        {li2023kbbinder}
\bibfield{author}{\bibinfo{person}{Tianle Li}, \bibinfo{person}{Xueguang Ma}, \bibinfo{person}{Alex Zhuang}, \bibinfo{person}{Yu Gu}, \bibinfo{person}{Yu Su}, {and} \bibinfo{person}{Wenhu Chen}.} \bibinfo{year}{2023}\natexlab{}.
\newblock \showarticletitle{{Few-shot In-context Learning on Knowledge Base Question Answering}}. In \bibinfo{booktitle}{\emph{ACL}}. \bibinfo{pages}{6966--6980}.
\newblock
\href{https://doi.org/10.18653/v1/2023.acl-long.385}{doi:\nolinkurl{10.18653/v1/2023.acl-long.385}}

\bibitem[Liu et~al\mbox{.}(2025b)]%
        {liu2025ort}
\bibfield{author}{\bibinfo{person}{Runxuan Liu}, \bibinfo{person}{Bei Luo}, \bibinfo{person}{Jiaqi Li}, \bibinfo{person}{Baoxin Wang}, \bibinfo{person}{Ming Liu}, \bibinfo{person}{Dayong Wu}, \bibinfo{person}{Shijin Wang}, {and} \bibinfo{person}{Bing Qin}.} \bibinfo{year}{2025}\natexlab{b}.
\newblock \showarticletitle{{Ontology-Guided Reverse Thinking Makes Large Language Models Stronger on Knowledge Graph Question Answering}}. In \bibinfo{booktitle}{\emph{ACL}}. \bibinfo{pages}{15269--15284}.
\newblock

\bibitem[Liu et~al\mbox{.}(2025a)]%
        {ontotune}
\bibfield{author}{\bibinfo{person}{Zhiqiang Liu}, \bibinfo{person}{Chengtao Gan}, \bibinfo{person}{Junjie Wang}, \bibinfo{person}{Yichi Zhang}, \bibinfo{person}{Zhongpu Bo}, \bibinfo{person}{Mengshu Sun}, \bibinfo{person}{Huajun Chen}, {and} \bibinfo{person}{Wen Zhang}.} \bibinfo{year}{2025}\natexlab{a}.
\newblock \showarticletitle{{OntoTune: Ontology-Driven Self-training for Aligning Large Language Models}}. In \bibinfo{booktitle}{\emph{WWW}}. \bibinfo{pages}{119--133}.
\newblock

\bibitem[Luo et~al\mbox{.}(2024a)]%
        {luo2024chatkbqa}
\bibfield{author}{\bibinfo{person}{Haoran Luo}, \bibinfo{person}{Haihong E}, \bibinfo{person}{Zichen Tang}, \bibinfo{person}{Shiyao Peng}, \bibinfo{person}{Yikai Guo}, \bibinfo{person}{Wentai Zhang}, \bibinfo{person}{Chenghao Ma}, \bibinfo{person}{Guanting Dong}, \bibinfo{person}{Meina Song}, \bibinfo{person}{Wei Lin}, \bibinfo{person}{Yifan Zhu}, {and} \bibinfo{person}{Anh~Tuan Luu}.} \bibinfo{year}{2024}\natexlab{a}.
\newblock \showarticletitle{{ChatKBQA: A Generate-then-Retrieve Framework for Knowledge Base Question Answering with Fine-tuned Large Language Models}}. In \bibinfo{booktitle}{\emph{Findings of ACL}}. \bibinfo{pages}{2039--2056}.
\newblock
\href{https://doi.org/10.18653/v1/2024.findings-acl.122}{doi:\nolinkurl{10.18653/v1/2024.findings-acl.122}}

\bibitem[Luo et~al\mbox{.}(2024b)]%
        {luo2024rog}
\bibfield{author}{\bibinfo{person}{Linhao Luo}, \bibinfo{person}{Yuan-Fang Li}, \bibinfo{person}{Gholamreza Haffari}, {and} \bibinfo{person}{Shirui Pan}.} \bibinfo{year}{2024}\natexlab{b}.
\newblock \showarticletitle{{Reasoning on Graphs: Faithful and Interpretable Large Language Model Reasoning}}. In \bibinfo{booktitle}{\emph{ICLR}}.
\newblock

\bibitem[Ma et~al\mbox{.}(2025)]%
        {ma2025tog2}
\bibfield{author}{\bibinfo{person}{Shengjie Ma}, \bibinfo{person}{Chengjin Xu}, \bibinfo{person}{Xuhui Jiang}, \bibinfo{person}{Muzhi Li}, \bibinfo{person}{Huaren Qu}, \bibinfo{person}{Cehao Yang}, \bibinfo{person}{Jiaxin Mao}, {and} \bibinfo{person}{Jian Guo}.} \bibinfo{year}{2025}\natexlab{}.
\newblock \showarticletitle{{Think-on-Graph 2.0: Deep and Faithful Large Language Model Reasoning with Knowledge-guided Retrieval Augmented Generation}}. In \bibinfo{booktitle}{\emph{ICLR}}.
\newblock

\bibitem[Moskvoretskii et~al\mbox{.}(2024)]%
        {moskvoretskii2024taxollama}
\bibfield{author}{\bibinfo{person}{Viktor Moskvoretskii}, \bibinfo{person}{Ekaterina Neminova}, \bibinfo{person}{Alina Lobanova}, \bibinfo{person}{Alexander Panchenko}, {and} \bibinfo{person}{Irina Nikishina}.} \bibinfo{year}{2024}\natexlab{}.
\newblock \showarticletitle{{TaxoLLaMA: WordNet-based Model for Solving Multiple Lexical Semantic Tasks}}. In \bibinfo{booktitle}{\emph{ACL}}. \bibinfo{pages}{2331--2350}.
\newblock

\bibitem[Muennighoff et~al\mbox{.}(2025)]%
        {muennighoff2025s1}
\bibfield{author}{\bibinfo{person}{Niklas Muennighoff}, \bibinfo{person}{Zitong Yang}, \bibinfo{person}{Weijia Shi}, \bibinfo{person}{Xiang~Lisa Li}, \bibinfo{person}{Li Fei-Fei}, \bibinfo{person}{Hannaneh Hajishirzi}, \bibinfo{person}{Luke Zettlemoyer}, \bibinfo{person}{Percy Liang}, \bibinfo{person}{Emmanuel Cand{\`e}s}, {and} \bibinfo{person}{Tatsunori Hashimoto}.} \bibinfo{year}{2025}\natexlab{}.
\newblock \showarticletitle{{s1: Simple Test-Time Scaling}}. In \bibinfo{booktitle}{\emph{EMNLP}}. \bibinfo{pages}{20275--20321}.
\newblock

\bibitem[Olausson et~al\mbox{.}(2023)]%
        {olausson2023linc}
\bibfield{author}{\bibinfo{person}{Theo Olausson}, \bibinfo{person}{Alex Gu}, \bibinfo{person}{Ben Lipkin}, \bibinfo{person}{Cedegao Zhang}, \bibinfo{person}{Armando Solar-Lezama}, \bibinfo{person}{Joshua Tenenbaum}, {and} \bibinfo{person}{Roger Levy}.} \bibinfo{year}{2023}\natexlab{}.
\newblock \showarticletitle{{LINC: A Neurosymbolic Approach for Logical Reasoning by Combining Language Models with First-Order Logic Provers}}. In \bibinfo{booktitle}{\emph{EMNLP}}. \bibinfo{pages}{5153--5176}.
\newblock
\href{https://doi.org/10.18653/v1/2023.emnlp-main.313}{doi:\nolinkurl{10.18653/v1/2023.emnlp-main.313}}

\bibitem[Pal et~al\mbox{.}(2022)]%
        {pal2022medmcqa}
\bibfield{author}{\bibinfo{person}{Ankit Pal}, \bibinfo{person}{Logesh~Kumar Umapathi}, {and} \bibinfo{person}{Malaikannan Sankarasubbu}.} \bibinfo{year}{2022}\natexlab{}.
\newblock \showarticletitle{{MedMCQA: A Large-scale Multi-Subject Multi-Choice Dataset for Medical Domain Question Answering}}. In \bibinfo{booktitle}{\emph{CHIL}}. \bibinfo{pages}{248--260}.
\newblock

\bibitem[Pan et~al\mbox{.}(2023)]%
        {pan2023logiclm}
\bibfield{author}{\bibinfo{person}{Liangming Pan}, \bibinfo{person}{Alon Albalak}, \bibinfo{person}{Xinyi Wang}, {and} \bibinfo{person}{William Wang}.} \bibinfo{year}{2023}\natexlab{}.
\newblock \showarticletitle{{Logic-LM: Empowering Large Language Models with Symbolic Solvers for Faithful Logical Reasoning}}. In \bibinfo{booktitle}{\emph{Findings of EMNLP}}. \bibinfo{pages}{3806--3824}.
\newblock

\bibitem[Pellissier~Tanon(2023)]%
        {oxigraph}
\bibfield{author}{\bibinfo{person}{Thomas Pellissier~Tanon}.} \bibinfo{year}{2023}\natexlab{}.
\newblock \bibinfo{title}{{Oxigraph: a graph database implementing the SPARQL standard}}.
\newblock
\href{https://doi.org/10.5281/zenodo.7408022}{doi:\nolinkurl{10.5281/zenodo.7408022}}
\newblock
\shownote{Software, \url{https://oxigraph.org}}.

\bibitem[Qin et~al\mbox{.}(2025)]%
        {qin2024toollearning}
\bibfield{author}{\bibinfo{person}{Yujia Qin}, \bibinfo{person}{Shengding Hu}, \bibinfo{person}{Yankai Lin}, \bibinfo{person}{Weize Chen}, \bibinfo{person}{Ning Ding}, \bibinfo{person}{Ganqu Cui}, \bibinfo{person}{Zheni Zeng}, \bibinfo{person}{Xuanhe Zhou}, \bibinfo{person}{Yufei Huang}, \bibinfo{person}{Chaojun Xiao}, {et~al\mbox{.}}} \bibinfo{year}{2025}\natexlab{}.
\newblock \showarticletitle{{Tool Learning with Foundation Models}}.
\newblock \bibinfo{journal}{\emph{Comput. Surveys}} \bibinfo{volume}{57}, \bibinfo{number}{4} (\bibinfo{year}{2025}), \bibinfo{pages}{1--40}.
\newblock
\href{https://doi.org/10.1145/3704435}{doi:\nolinkurl{10.1145/3704435}}

\bibitem[{Qwen Team}(2026)]%
        {qwen35}
\bibfield{author}{\bibinfo{person}{{Qwen Team}}.} \bibinfo{year}{2026}\natexlab{}.
\newblock \bibinfo{title}{{Qwen3.5: Towards Native Multimodal Agents}}.
\newblock \bibinfo{howpublished}{\url{https://qwen.ai/blog?id=qwen3.5}}.
\newblock

\bibitem[Ratcliff and Metzener(1988)]%
        {ratcliff1988gestalt}
\bibfield{author}{\bibinfo{person}{John~W. Ratcliff} {and} \bibinfo{person}{David~E. Metzener}.} \bibinfo{year}{1988}\natexlab{}.
\newblock \showarticletitle{Pattern Matching: The Gestalt Approach}.
\newblock \bibinfo{journal}{\emph{Dr. Dobb's Journal}} \bibinfo{volume}{13}, \bibinfo{number}{7} (\bibinfo{year}{1988}), \bibinfo{pages}{46}.
\newblock

\bibitem[Reimers and Gurevych(2019)]%
        {sbert}
\bibfield{author}{\bibinfo{person}{Nils Reimers} {and} \bibinfo{person}{Iryna Gurevych}.} \bibinfo{year}{2019}\natexlab{}.
\newblock \showarticletitle{{Sentence-BERT: Sentence Embeddings using Siamese BERT-Networks}}. In \bibinfo{booktitle}{\emph{EMNLP-IJCNLP}}. \bibinfo{pages}{3982--3992}.
\newblock

\bibitem[Robertson and Zaragoza(2009)]%
        {bm25}
\bibfield{author}{\bibinfo{person}{Stephen Robertson} {and} \bibinfo{person}{Hugo Zaragoza}.} \bibinfo{year}{2009}\natexlab{}.
\newblock \showarticletitle{{The Probabilistic Relevance Framework: BM25 and Beyond}}.
\newblock \bibinfo{journal}{\emph{Foundations and Trends in Information Retrieval}} \bibinfo{volume}{3}, \bibinfo{number}{4} (\bibinfo{year}{2009}), \bibinfo{pages}{333--389}.
\newblock
\href{https://doi.org/10.1561/1500000019}{doi:\nolinkurl{10.1561/1500000019}}

\bibitem[Schick et~al\mbox{.}(2023)]%
        {schick2023toolformer}
\bibfield{author}{\bibinfo{person}{Timo Schick}, \bibinfo{person}{Jane Dwivedi-Yu}, \bibinfo{person}{Roberto Dess{\`i}}, \bibinfo{person}{Roberta Raileanu}, \bibinfo{person}{Maria Lomeli}, \bibinfo{person}{Eric Hambro}, \bibinfo{person}{Luke Zettlemoyer}, \bibinfo{person}{Nicola Cancedda}, {and} \bibinfo{person}{Thomas Scialom}.} \bibinfo{year}{2023}\natexlab{}.
\newblock \showarticletitle{{Toolformer: Language Models Can Teach Themselves to Use Tools}}. In \bibinfo{booktitle}{\emph{NeurIPS}}.
\newblock

\bibitem[Sharma et~al\mbox{.}(2025)]%
        {sharma2025ograg}
\bibfield{author}{\bibinfo{person}{Kartik Sharma}, \bibinfo{person}{Peeyush Kumar}, {and} \bibinfo{person}{Yunqing Li}.} \bibinfo{year}{2025}\natexlab{}.
\newblock \showarticletitle{{OG-RAG: Ontology-Grounded Retrieval-Augmented Generation for Large Language Models}}. In \bibinfo{booktitle}{\emph{EMNLP}}. \bibinfo{pages}{32962--32981}.
\newblock
\href{https://doi.org/10.18653/v1/2025.emnlp-main.1674}{doi:\nolinkurl{10.18653/v1/2025.emnlp-main.1674}}

\bibitem[Smith et~al\mbox{.}(2007)]%
        {smith2007obo}
\bibfield{author}{\bibinfo{person}{Barry Smith}, \bibinfo{person}{Michael Ashburner}, \bibinfo{person}{Cornelius Rosse}, \bibinfo{person}{Jonathan Bard}, \bibinfo{person}{William Bug}, \bibinfo{person}{Werner Ceusters}, \bibinfo{person}{Louis~J. Goldberg}, \bibinfo{person}{Karen Eilbeck}, \bibinfo{person}{Amelia Ireland}, \bibinfo{person}{Christopher~J. Mungall}, {et~al\mbox{.}}} \bibinfo{year}{2007}\natexlab{}.
\newblock \showarticletitle{{The OBO Foundry: Coordinated Evolution of Ontologies to Support Biomedical Data Integration}}.
\newblock \bibinfo{journal}{\emph{Nature Biotechnology}} \bibinfo{volume}{25}, \bibinfo{number}{11} (\bibinfo{year}{2007}), \bibinfo{pages}{1251--1255}.
\newblock
\href{https://doi.org/10.1038/nbt1346}{doi:\nolinkurl{10.1038/nbt1346}}

\bibitem[{StatPearls Publishing}(2026)]%
        {statpearls}
\bibfield{author}{\bibinfo{person}{{StatPearls Publishing}}.} \bibinfo{year}{2026}\natexlab{}.
\newblock \bibinfo{title}{{StatPearls}}.
\newblock \bibinfo{howpublished}{NCBI Bookshelf NBK430685, \url{https://www.ncbi.nlm.nih.gov/books/NBK430685/}}.
\newblock
\newblock
\shownote{Archive of 27 September 2026}.

\bibitem[Su et~al\mbox{.}(2025)]%
        {su2025kgarevion}
\bibfield{author}{\bibinfo{person}{Xiaorui Su}, \bibinfo{person}{Yibo Wang}, \bibinfo{person}{Shanghua Gao}, \bibinfo{person}{Xiaolong Liu}, \bibinfo{person}{Valentina Giunchiglia}, \bibinfo{person}{Djork-Arn{\'e} Clevert}, {and} \bibinfo{person}{Marinka Zitnik}.} \bibinfo{year}{2025}\natexlab{}.
\newblock \showarticletitle{{KGARevion: An AI Agent for Knowledge-Intensive Biomedical QA}}. In \bibinfo{booktitle}{\emph{ICLR}}.
\newblock

\bibitem[Sun et~al\mbox{.}(2024)]%
        {sun2024tog}
\bibfield{author}{\bibinfo{person}{Jiashuo Sun}, \bibinfo{person}{Chengjin Xu}, \bibinfo{person}{Lumingyuan Tang}, \bibinfo{person}{Saizhuo Wang}, \bibinfo{person}{Chen Lin}, \bibinfo{person}{Yeyun Gong}, \bibinfo{person}{Lionel~M. Ni}, \bibinfo{person}{Heung-Yeung Shum}, {and} \bibinfo{person}{Jian Guo}.} \bibinfo{year}{2024}\natexlab{}.
\newblock \showarticletitle{{Think-on-Graph: Deep and Responsible Reasoning of Large Language Model on Knowledge Graph}}. In \bibinfo{booktitle}{\emph{ICLR}}.
\newblock

\bibitem[Trivedi et~al\mbox{.}(2023)]%
        {trivedi2023ircot}
\bibfield{author}{\bibinfo{person}{Harsh Trivedi}, \bibinfo{person}{Niranjan Balasubramanian}, \bibinfo{person}{Tushar Khot}, {and} \bibinfo{person}{Ashish Sabharwal}.} \bibinfo{year}{2023}\natexlab{}.
\newblock \showarticletitle{{Interleaving Retrieval with Chain-of-Thought Reasoning for Knowledge-Intensive Multi-Step Questions}}. In \bibinfo{booktitle}{\emph{ACL}}. \bibinfo{pages}{10014--10037}.
\newblock

\bibitem[Wang et~al\mbox{.}(2020)]%
        {wang2020minilm}
\bibfield{author}{\bibinfo{person}{Wenhui Wang}, \bibinfo{person}{Furu Wei}, \bibinfo{person}{Li Dong}, \bibinfo{person}{Hangbo Bao}, \bibinfo{person}{Nan Yang}, {and} \bibinfo{person}{Ming Zhou}.} \bibinfo{year}{2020}\natexlab{}.
\newblock \showarticletitle{{MiniLM: Deep Self-Attention Distillation for Task-Agnostic Compression of Pre-Trained Transformers}}. In \bibinfo{booktitle}{\emph{NeurIPS}}, Vol.~\bibinfo{volume}{33}. \bibinfo{pages}{5776--5788}.
\newblock

\bibitem[Wang et~al\mbox{.}(2024)]%
        {wang2024chainoftable}
\bibfield{author}{\bibinfo{person}{Zilong Wang}, \bibinfo{person}{Hao Zhang}, \bibinfo{person}{Chun-Liang Li}, \bibinfo{person}{Julian~Martin Eisenschlos}, \bibinfo{person}{Vincent Perot}, \bibinfo{person}{Zifeng Wang}, \bibinfo{person}{Lesly Miculicich}, \bibinfo{person}{Yasuhisa Fujii}, \bibinfo{person}{Jingbo Shang}, \bibinfo{person}{Chen-Yu Lee}, {and} \bibinfo{person}{Tomas Pfister}.} \bibinfo{year}{2024}\natexlab{}.
\newblock \showarticletitle{{Chain-of-Table: Evolving Tables in the Reasoning Chain for Table Understanding}}. In \bibinfo{booktitle}{\emph{ICLR}}.
\newblock

\bibitem[Wei et~al\mbox{.}(2022)]%
        {wei2022cot}
\bibfield{author}{\bibinfo{person}{Jason Wei}, \bibinfo{person}{Xuezhi Wang}, \bibinfo{person}{Dale Schuurmans}, \bibinfo{person}{Maarten Bosma}, \bibinfo{person}{Brian Ichter}, \bibinfo{person}{Fei Xia}, \bibinfo{person}{Ed Chi}, \bibinfo{person}{Quoc~V. Le}, {and} \bibinfo{person}{Denny Zhou}.} \bibinfo{year}{2022}\natexlab{}.
\newblock \showarticletitle{{Chain-of-Thought Prompting Elicits Reasoning in Large Language Models}}. In \bibinfo{booktitle}{\emph{NeurIPS}}.
\newblock

\bibitem[Wu et~al\mbox{.}(2025)]%
        {wu2025medgraphrag}
\bibfield{author}{\bibinfo{person}{Junde Wu}, \bibinfo{person}{Jiayuan Zhu}, \bibinfo{person}{Yunli Qi}, \bibinfo{person}{Jingkun Chen}, \bibinfo{person}{Min Xu}, \bibinfo{person}{Filippo Menolascina}, \bibinfo{person}{Yueming Jin}, {and} \bibinfo{person}{Vicente Grau}.} \bibinfo{year}{2025}\natexlab{}.
\newblock \showarticletitle{{Medical Graph RAG: Evidence-based Medical Large Language Model via Graph Retrieval-Augmented Generation}}. In \bibinfo{booktitle}{\emph{ACL}}. \bibinfo{pages}{28443--28467}.
\newblock

\bibitem[Xiong et~al\mbox{.}(2024a)]%
        {xiong2024medrag}
\bibfield{author}{\bibinfo{person}{Guangzhi Xiong}, \bibinfo{person}{Qiao Jin}, \bibinfo{person}{Zhiyong Lu}, {and} \bibinfo{person}{Aidong Zhang}.} \bibinfo{year}{2024}\natexlab{a}.
\newblock \showarticletitle{{Benchmarking Retrieval-Augmented Generation for Medicine}}. In \bibinfo{booktitle}{\emph{Findings of ACL}}. \bibinfo{pages}{6233--6251}.
\newblock

\bibitem[Xiong et~al\mbox{.}(2024b)]%
        {xiong2024imedrag}
\bibfield{author}{\bibinfo{person}{Guangzhi Xiong}, \bibinfo{person}{Qiao Jin}, \bibinfo{person}{Xiao Wang}, \bibinfo{person}{Minjia Zhang}, \bibinfo{person}{Zhiyong Lu}, {and} \bibinfo{person}{Aidong Zhang}.} \bibinfo{year}{2024}\natexlab{b}.
\newblock \showarticletitle{{Improving Retrieval-Augmented Generation in Medicine with Iterative Follow-up Questions}}. In \bibinfo{booktitle}{\emph{PSB}}. \bibinfo{pages}{199--214}.
\newblock
\href{https://doi.org/10.1142/9789819807024_0015}{doi:\nolinkurl{10.1142/9789819807024_0015}}

\bibitem[Yang et~al\mbox{.}(2025)]%
        {yang2025ont}
\bibfield{author}{\bibinfo{person}{Hui Yang}, \bibinfo{person}{Jiaoyan Chen}, \bibinfo{person}{Yuan He}, \bibinfo{person}{Yongsheng Gao}, {and} \bibinfo{person}{Ian Horrocks}.} \bibinfo{year}{2025}\natexlab{}.
\newblock \showarticletitle{{Language Models as Ontology Encoders}}. In \bibinfo{booktitle}{\emph{ISWC}}. \bibinfo{pages}{443--461}.
\newblock

\bibitem[Yang et~al\mbox{.}(2026)]%
        {yang2026neurowl}
\bibfield{author}{\bibinfo{person}{Hui Yang}, \bibinfo{person}{Jiaoyan Chen}, \bibinfo{person}{Yiping Song}, \bibinfo{person}{Renate Schmidt}, {and} \bibinfo{person}{Wen Zhang}.} \bibinfo{year}{2026}\natexlab{}.
\newblock \bibinfo{title}{{NeurOWL: An LLM-Based Neural-symbolic Framework for Incomplete OWL Ontology Reasoning}}.
\newblock
\newblock
\shownote{arXiv:2607.15776}.

\bibitem[Yang et~al\mbox{.}(2024)]%
        {yang2024kgrank}
\bibfield{author}{\bibinfo{person}{Rui Yang}, \bibinfo{person}{Haoran Liu}, \bibinfo{person}{Edison Marrese-Taylor}, \bibinfo{person}{Qingcheng Zeng}, \bibinfo{person}{Yuhe Ke}, \bibinfo{person}{Wanxin Li}, \bibinfo{person}{Lechao Cheng}, \bibinfo{person}{Qingyu Chen}, \bibinfo{person}{James Caverlee}, \bibinfo{person}{Yutaka Matsuo}, {and} \bibinfo{person}{Irene Li}.} \bibinfo{year}{2024}\natexlab{}.
\newblock \showarticletitle{{KG-Rank: Enhancing Large Language Models for Medical QA with Knowledge Graphs and Ranking Techniques}}. In \bibinfo{booktitle}{\emph{BioNLP Workshop}}. \bibinfo{pages}{155--166}.
\newblock

\bibitem[Yao et~al\mbox{.}(2023)]%
        {yao2023react}
\bibfield{author}{\bibinfo{person}{Shunyu Yao}, \bibinfo{person}{Jeffrey Zhao}, \bibinfo{person}{Dian Yu}, \bibinfo{person}{Nan Du}, \bibinfo{person}{Izhak Shafran}, \bibinfo{person}{Karthik Narasimhan}, {and} \bibinfo{person}{Yuan Cao}.} \bibinfo{year}{2023}\natexlab{}.
\newblock \showarticletitle{{ReAct: Synergizing Reasoning and Acting in Language Models}}. In \bibinfo{booktitle}{\emph{ICLR}}.
\newblock

\bibitem[Zhang et~al\mbox{.}(2026)]%
        {zhang2025ontourl}
\bibfield{author}{\bibinfo{person}{Xiao Zhang}, \bibinfo{person}{Huiyuan Lai}, \bibinfo{person}{Qianru Meng}, {and} \bibinfo{person}{Johan Bos}.} \bibinfo{year}{2026}\natexlab{}.
\newblock \showarticletitle{{OntoURL: A Benchmark for Evaluating Large Language Models on Symbolic Ontological Understanding, Reasoning and Learning}}.
\newblock \bibinfo{journal}{\emph{Journal of Web Semantics}}  \bibinfo{volume}{91} (\bibinfo{year}{2026}), \bibinfo{pages}{100886}.
\newblock
\href{https://doi.org/10.1016/j.websem.2026.100886}{doi:\nolinkurl{10.1016/j.websem.2026.100886}}

\end{thebibliography}

\appendix

\section{{Implementation Details}}
\label{app:impl}

\textbf{Agent and service.} For each item, \ours{} allows up to 10 tool actions and 5,000 generated tokens across pre-tool reasoning, the ReAct loop, and Step~1 of answering, plus 512 tokens for Step~2. The ontology service decides \code{ASK} and \code{SELECT} queries over the entailment closure of the ontology, precomputed with the OWL EL reasoner Whelk~\cite{balhoff2024whelk} and stored in Oxigraph~\cite{oxigraph}; the datasets are built and checked with ELK~\cite{kazakov2014elk} and HermiT~\cite{glimm2014hermit}.

\textbf{Cost.} On Test~3 with Qwen, \ours{} uses 9.7k tokens per question over 5.8 LLM calls on average, including 7.9k input tokens (1.0k from ontology observations) and 1.8k generated tokens. In contrast, i-MedRAG$_{\texttt{doc}}$\fwmark{} uses 96.7k tokens over 22.0 calls, including 87.8k input tokens (63.6k from retrieved documents) and 8.9k generated tokens.

\begin{figure}[b]
\centering
\scalebox{0.8}{\begin{tikzpicture}[x=1cm, y=1cm, font=\small]
\path (0,0.02) -- (8.40,0.02);
\node[draw=black!40, line width=0.4pt, rounded corners=2pt, inner xsep=3pt, inner ysep=2.5pt, anchor=north] at (4.200,0) {\setlength{\tabcolsep}{0pt}\begin{tabular}{@{}l@{\hspace{10pt}}l@{\hspace{10pt}}l@{}}\tikz[baseline=-0.6ex]\filldraw[fill=fDesc, draw=black!40, line width=0.3pt, sharp corners] (0,-0.085) rectangle (0.30,0.085);\,DESCRIBE & \tikz[baseline=-0.6ex]\filldraw[fill=fAsk, draw=black!40, line width=0.3pt, sharp corners] (0,-0.085) rectangle (0.30,0.085);\,ASK & \tikz[baseline=-0.6ex]\filldraw[fill=fSelx, draw=black!40, line width=0.3pt, sharp corners] (0,-0.085) rectangle (0.30,0.085);\,SELECT\end{tabular}};
\node[rotate=90, anchor=center, inner sep=0pt] at (0.160,-1.650) {Llama};
\draw[black!10, line width=0.4pt] (2.000,-0.750) -- (2.000,-2.550);
\draw[black!10, line width=0.4pt] (3.729,-0.750) -- (3.729,-2.550);
\draw[black!10, line width=0.4pt] (5.457,-0.750) -- (5.457,-2.550);
\draw[black!10, line width=0.4pt] (7.186,-0.750) -- (7.186,-2.550);
\node[rotate=90, anchor=center, inner sep=0pt] at (0.160,-3.630) {Qwen};
\draw[black!10, line width=0.4pt] (2.000,-2.730) -- (2.000,-4.550);
\draw[black!10, line width=0.4pt] (3.729,-2.730) -- (3.729,-4.550);
\draw[black!10, line width=0.4pt] (5.457,-2.730) -- (5.457,-4.550);
\draw[black!10, line width=0.4pt] (7.186,-2.730) -- (7.186,-4.550);
\draw[black!25, line width=0.4pt] (0,-2.640) -- (8.400,-2.640);
\node[anchor=east, inner sep=0pt, text height=1.5ex, text depth=0.25ex] at (1.930,-0.930) {Test 1 MCQ};
\path[fill=fDesc, draw=white, line width=0.45pt] (2.000,-0.800) rectangle (2.337,-1.060);
\path[fill=fAsk, draw=white, line width=0.45pt] (2.337,-0.800) rectangle (3.724,-1.060);
\path[fill=fSelx, draw=white, line width=0.45pt] (3.724,-0.800) rectangle (5.100,-1.060);
\node[anchor=west, inner sep=0pt, inner sep=0pt, text height=1.5ex, text depth=0.25ex] at (5.160,-0.930) {3.6};
\node[anchor=east, inner sep=0pt, text height=1.5ex, text depth=0.25ex] at (1.930,-1.290) {Test 1 OQ};
\path[fill=fDesc, draw=white, line width=0.45pt] (2.000,-1.160) rectangle (2.073,-1.420);
\path[fill=fAsk, draw=white, line width=0.45pt] (2.073,-1.160) rectangle (2.094,-1.420);
\path[fill=fSelx, draw=white, line width=0.45pt] (2.094,-1.160) rectangle (5.323,-1.420);
\node[anchor=west, inner sep=0pt, inner sep=0pt, text height=1.5ex, text depth=0.25ex] at (5.383,-1.290) {3.8};
\node[anchor=east, inner sep=0pt, text height=1.5ex, text depth=0.25ex] at (1.930,-1.650) {Test 2 MCQ};
\path[fill=fDesc, draw=white, line width=0.45pt] (2.000,-1.520) rectangle (2.637,-1.780);
\path[fill=fAsk, draw=white, line width=0.45pt] (2.637,-1.520) rectangle (4.282,-1.780);
\path[fill=fSelx, draw=white, line width=0.45pt] (4.282,-1.520) rectangle (6.114,-1.780);
\node[anchor=west, inner sep=0pt, inner sep=0pt, text height=1.5ex, text depth=0.25ex] at (6.174,-1.650) {4.8};
\node[anchor=east, inner sep=0pt, text height=1.5ex, text depth=0.25ex] at (1.930,-2.010) {Test 2 OQ};
\path[fill=fDesc, draw=white, line width=0.45pt] (2.000,-1.880) rectangle (2.107,-2.140);
\path[fill=fAsk, draw=white, line width=0.45pt] (2.107,-1.880) rectangle (2.114,-2.140);
\path[fill=fSelx, draw=white, line width=0.45pt] (2.114,-1.880) rectangle (6.249,-2.140);
\node[anchor=west, inner sep=0pt, inner sep=0pt, text height=1.5ex, text depth=0.25ex] at (6.309,-2.010) {4.9};
\node[anchor=east, inner sep=0pt, text height=1.5ex, text depth=0.25ex] at (1.930,-2.370) {Test 3};
\path[fill=fDesc, draw=white, line width=0.45pt] (2.000,-2.240) rectangle (2.773,-2.500);
\path[fill=fAsk, draw=white, line width=0.45pt] (2.773,-2.240) rectangle (4.491,-2.500);
\path[fill=fSelx, draw=white, line width=0.45pt] (4.491,-2.240) rectangle (6.256,-2.500);
\node[anchor=west, inner sep=0pt, inner sep=0pt, text height=1.5ex, text depth=0.25ex] at (6.316,-2.370) {4.9};
\node[anchor=east, inner sep=0pt, text height=1.5ex, text depth=0.25ex] at (1.930,-2.910) {Test 1 MCQ};
\path[fill=fDesc, draw=white, line width=0.45pt] (2.000,-2.780) rectangle (2.935,-3.040);
\path[fill=fAsk, draw=white, line width=0.45pt] (2.935,-2.780) rectangle (3.246,-3.040);
\path[fill=fSelx, draw=white, line width=0.45pt] (3.246,-2.780) rectangle (3.926,-3.040);
\node[anchor=west, inner sep=0pt, inner sep=0pt, text height=1.5ex, text depth=0.25ex] at (3.986,-2.910) {2.2};
\node[anchor=east, inner sep=0pt, text height=1.5ex, text depth=0.25ex] at (1.930,-3.270) {Test 1 OQ};
\path[fill=fDesc, draw=white, line width=0.45pt] (2.000,-3.140) rectangle (4.155,-3.400);
\path[fill=fAsk, draw=white, line width=0.45pt] (4.155,-3.140) rectangle (4.358,-3.400);
\path[fill=fSelx, draw=white, line width=0.45pt] (4.358,-3.140) rectangle (5.875,-3.400);
\node[anchor=west, inner sep=0pt, inner sep=0pt, text height=1.5ex, text depth=0.25ex] at (5.935,-3.270) {4.5};
\node[anchor=east, inner sep=0pt, text height=1.5ex, text depth=0.25ex] at (1.930,-3.630) {Test 2 MCQ};
\path[fill=fDesc, draw=white, line width=0.45pt] (2.000,-3.500) rectangle (3.498,-3.760);
\path[fill=fAsk, draw=white, line width=0.45pt] (3.498,-3.500) rectangle (3.904,-3.760);
\path[fill=fSelx, draw=white, line width=0.45pt] (3.904,-3.500) rectangle (4.813,-3.760);
\node[anchor=west, inner sep=0pt, inner sep=0pt, text height=1.5ex, text depth=0.25ex] at (4.873,-3.630) {3.3};
\node[anchor=east, inner sep=0pt, text height=1.5ex, text depth=0.25ex] at (1.930,-3.990) {Test 2 OQ};
\path[fill=fDesc, draw=white, line width=0.45pt] (2.000,-3.860) rectangle (4.938,-4.120);
\path[fill=fAsk, draw=white, line width=0.45pt] (4.938,-3.860) rectangle (5.207,-4.120);
\path[fill=fSelx, draw=white, line width=0.45pt] (5.207,-3.860) rectangle (7.254,-4.120);
\node[anchor=west, inner sep=0pt, inner sep=0pt, text height=1.5ex, text depth=0.25ex] at (7.314,-3.990) {6.1};
\node[anchor=east, inner sep=0pt, text height=1.5ex, text depth=0.25ex] at (1.930,-4.350) {Test 3};
\path[fill=fDesc, draw=white, line width=0.45pt] (2.000,-4.220) rectangle (3.484,-4.480);
\path[fill=fAsk, draw=white, line width=0.45pt] (3.484,-4.220) rectangle (3.594,-4.480);
\path[fill=fSelx, draw=white, line width=0.45pt] (3.594,-4.220) rectangle (3.759,-4.480);
\node[anchor=west, inner sep=0pt, inner sep=0pt, text height=1.5ex, text depth=0.25ex] at (3.819,-4.350) {2.0};
\draw[black!45, line width=0.4pt] (2.000,-4.550) -- (8.050,-4.550);
\node[anchor=north, inner sep=1.5pt] at (2.000,-4.550) {0};
\node[anchor=north, inner sep=1.5pt] at (3.729,-4.550) {2};
\node[anchor=north, inner sep=1.5pt] at (5.457,-4.550) {4};
\node[anchor=north, inner sep=1.5pt] at (7.186,-4.550) {6};
\end{tikzpicture}}%
\caption{Tool calls of \ours{} per question by query form (\code{SELECT}: all its variants). Test~3 pools its six sets.}
\label{fig:calls-a}
\end{figure}
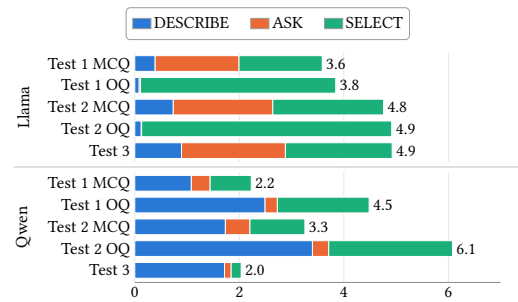

\textbf{Tool calls} (Figure~\ref{fig:calls-a}). Query choice varies systematically with question type and model. On the OQs of Tests~1 and~2, Llama relies almost exclusively on \code{SELECT} (97\% of calls), whereas Qwen primarily uses \code{DESCRIBE} (56\%). On the MCQs, Llama divides its calls between \code{ASK} and \code{SELECT} (40--45\% each), while Qwen again favors \code{DESCRIBE} (49--53\%). On Test~3, Llama averages 4.9 calls per question, split evenly between \code{ASK} and \code{SELECT} (40 and 42\%), whereas Qwen averages 2.0 calls and uses \code{DESCRIBE} for 84\% of them.

\section{{Retrieval and Document Baselines}}
\label{app:retrieval}\label{app:docrag}

The retrieval baselines differ primarily in their knowledge source: RAG$_{\texttt{ont}}$(text), ReAct$_{\texttt{ont}}$, and the ablation \code{search} tool retrieve ontology concept cards, whereas RAG$_{\texttt{doc}}$(text), MedRAG$_{\texttt{doc}}$, and i-MedRAG$_{\texttt{doc}}$ retrieve medical documents.

\textbf{Ontology retrieval.} Each field of a concept card (\S\ref{sec:docs}) forms a retrieval unit, grouped by type (names, definitions, etc.). Units are retrieved using BM25~\cite{bm25} and sentence embeddings (\code{all-MiniLM-L6-v2}~\cite{sbert,wang2020minilm}), with the resulting rankings combined by reciprocal rank fusion~\cite{rrf}; each concept is scored according to its highest-ranked unit. RAG$_{\texttt{ont}}$(text) performs a single retrieval using the question and answer options, then inserts the top-5 concept cards together with the top-5 verbalized axioms retrieved using the same scoring procedure.
ReAct$_{\texttt{ont}}$ uses the same underlying retrieval mechanism as RAG$_{\texttt{ont}}$(text) but each search returns the top-5 concepts, each with only the field that matched rather than its complete concept card, followed by the top-5 matching verbalized axioms. Unlike RAG$_{\texttt{ont}}$(text), ReAct$_{\texttt{ont}}$ generates search queries dynamically using the LLM. The \code{search} tool used in our ablation study follows the same approach.

\textbf{Document baselines.} Following i-MedRAG, we use MedCPT~\cite{jin2023medcpt} to retrieve from medical textbooks~\cite{jin2021medqa} and StatPearls~\cite{statpearls}. RAG$_{\texttt{doc}}$(text) replaces ontology cards with the top-5 retrieved snippets. Under their standard settings, MedRAG$_{\texttt{doc}}$~\cite{xiong2024medrag} answers from 32 snippets in a single call, while i-MedRAG$_{\texttt{doc}}$~\cite{xiong2024imedrag} issues three follow-up queries in each of four rounds. Both are run with the authors' toolkit using a 7,168-token document budget, with a forced answer when the response reaches the context-window limit~\cite{muennighoff2025s1}.

\begin{table*}[t]
\centering
\caption{Test~1 and Test~2 accuracy (\%) by source ontology. \tfirst{Red} / \tsecond{blue}: best / second best per model and test.}
\label{tab:docrag-onto}
\renewcommand{\arraystretch}{0.8}%
\small
\setlength{\tabcolsep}{5.5pt}
\begin{tabular}{@{}l*{12}{c}@{}}
\toprule
\multirow{3}{*}{\textbf{Method}} & \multicolumn{6}{c}{\textbf{Test~1}} & \multicolumn{6}{c}{\textbf{Test~2}} \\
\cmidrule(lr){2-7}\cmidrule(lr){8-13}
 & \multicolumn{2}{c}{GO} & \multicolumn{2}{c}{FoodOn} & \multicolumn{2}{c}{SNOMED CT} & \multicolumn{2}{c}{GO} & \multicolumn{2}{c}{FoodOn} & \multicolumn{2}{c}{SNOMED CT} \\
\cmidrule(lr){2-3}\cmidrule(lr){4-5}\cmidrule(lr){6-7}\cmidrule(lr){8-9}\cmidrule(lr){10-11}\cmidrule(lr){12-13}
 & MCQ & OQ & MCQ & OQ & MCQ & OQ & MCQ & OQ & MCQ & OQ & MCQ & OQ \\
\midrule
\rowcolor{black!6}\multicolumn{13}{c}{\textit{Llama}} \\
Direct & 50.7 & 5.0 & 33.7 & \tsecond{12.0} & \tsecond{51.0} & 6.5 & 50.7 & 5.0 & 33.8 & 12.4 & 51.9 & 6.9 \\
CoT & 47.7 & 1.0 & 32.0 & 6.5 & 35.0 & 1.5 & 47.7 & 1.0 & 32.1 & 6.7 & 35.6 & 1.6 \\
RAG$_{\texttt{ont}}$(text) & \tsecond{55.0} & 0.5 & \tfirst{53.3} & 3.5 & 44.3 & 5.0 & \tfirst{55.7} & 1.5 & \tfirst{51.8} & 5.7 & \tsecond{52.5} & 6.9 \\
RAG$_{\texttt{ont}}$(query) & 52.7 & \tsecond{13.0} & 41.0 & 4.0 & 42.3 & \tsecond{13.0} & 48.3 & 1.5 & 39.8 & 3.1 & 39.7 & 2.6 \\
RAG$_{\texttt{doc}}$(text) & 50.0 & 0.0 & 34.0 & 4.5 & 45.0 & 4.0 & 50.0 & 0.0 & 34.1 & 4.6 & 45.8 & 4.2 \\
MedRAG$_{\texttt{doc}}$ & 50.7 & 2.5 & 31.7 & 6.5 & 37.7 & 8.0 & 50.7 & 2.5 & 31.8 & 6.7 & 38.3 & 8.5 \\
ReAct$_{\texttt{ont}}$ & 50.0 & 10.0 & \tsecond{44.7} & 8.0 & 40.3 & 11.5 & \tsecond{54.3} & \tfirst{13.0} & 43.8 & \tsecond{13.4} & 46.1 & \tsecond{11.1} \\
i-MedRAG$_{\texttt{doc}}$ & 46.7 & 0.0 & 31.7 & 1.0 & 37.7 & 0.5 & 46.7 & 0.0 & 31.8 & 1.0 & 38.0 & 0.5 \\
\ours{} & \tfirst{63.0} & \tfirst{52.0} & \tfirst{53.3} & \tfirst{18.5} & \tfirst{69.7} & \tfirst{64.5} & 53.7 & \tsecond{12.5} & \tsecond{46.8} & \tfirst{17.0} & \tfirst{56.3} & \tfirst{24.9} \\
\midrule
\rowcolor{black!6}\multicolumn{13}{c}{\textit{Qwen}} \\
Direct & 56.7 & 4.5 & 51.7 & 14.0 & 61.0 & 6.5 & 56.7 & 4.5 & 51.8 & 14.4 & 61.7 & 6.9 \\
CoT & 65.3 & 13.0 & 57.0 & 12.0 & 61.0 & 13.0 & 65.3 & 13.0 & 57.2 & 12.4 & 61.7 & 13.8 \\
RAG$_{\texttt{ont}}$(text) & 71.7 & 27.0 & \tsecond{71.7} & 15.0 & 62.7 & 12.5 & \tsecond{73.0} & \tsecond{30.0} & \tsecond{67.6} & 12.4 & 66.8 & 17.5 \\
RAG$_{\texttt{ont}}$(query) & 70.3 & 31.0 & 61.0 & 11.0 & \tsecond{73.3} & \tsecond{36.5} & 60.0 & 4.0 & 53.8 & 8.2 & 56.9 & 5.3 \\
RAG$_{\texttt{doc}}$(text) & 63.3 & 6.5 & 59.7 & 5.5 & 60.0 & 4.0 & 63.3 & 6.5 & 59.9 & 5.7 & 60.7 & 3.7 \\
MedRAG$_{\texttt{doc}}$ & 61.0 & 0.0 & 56.0 & 0.0 & 59.7 & 1.5 & 61.0 & 0.0 & 56.2 & 0.0 & 60.3 & 1.6 \\
ReAct$_{\texttt{ont}}$ & \tsecond{72.7} & \tsecond{39.0} & 71.3 & \tsecond{32.0} & 63.7 & 23.5 & 66.7 & \tfirst{42.0} & 61.5 & \tfirst{30.9} & \tsecond{67.8} & \tsecond{25.4} \\
i-MedRAG$_{\texttt{doc}}$ & 47.3 & 0.5 & 29.7 & 0.0 & 43.3 & 1.0 & 47.3 & 0.5 & 29.8 & 0.0 & 44.1 & 1.1 \\
i-MedRAG$_{\texttt{doc}}$\fwmark & 51.0 & 1.0 & 27.3 & 0.0 & 44.3 & 2.0 & 51.0 & 1.0 & 27.4 & 0.0 & 45.1 & 2.1 \\
\ours{} & \tfirst{79.7} & \tfirst{62.5} & \tfirst{78.3} & \tfirst{34.0} & \tfirst{81.7} & \tfirst{89.5} & \tfirst{74.0} & 22.0 & \tfirst{74.6} & \tsecond{21.1} & \tfirst{72.5} & \tfirst{26.5} \\
\bottomrule
\end{tabular}
\end{table*}

\textbf{Results by source ontology} (Table~\ref{tab:docrag-onto}). \ours{} achieves the best performance in 19 of the 24 settings, one of them tied. In the other five, it falls behind RAG$_{\texttt{ont}}$(text) (with Llama on the Test~2 GO and FoodOn MCQs, 53.7 vs.\ 55.7\% and 46.8 vs.\ 51.8\%) or ReAct$_{\texttt{ont}}$ (on three Test~2 OQ settings); the largest gap is on the Test~2 GO OQs with Qwen (22.0 vs.\ 42.0\%), mainly because Qwen's exact queries rarely retrieve a correct concept there (for 15 of the 200 questions, vs.\ 106 on Test~1), so it gives no answer to 62.5\% of them.

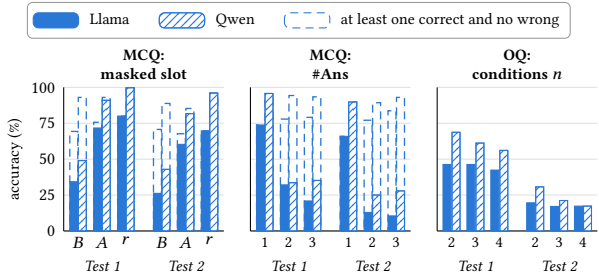
\begin{figure}[t]
\centering
\definecolor{vizthink}{HTML}{2A78D6}
\ref{strataleg}\par
\begin{tikzpicture}
\begin{groupplot}[group style={group size=3 by 1, horizontal sep=10pt, yticklabels at=edge left},
  width=0.25\columnwidth, height=1.9cm, ybar=0.4pt, /pgf/bar width=3pt, ymin=0, ymax=100, ytick={0,25,50,75,100},
  xtick={1,2,3,4.5,5.5,6.5}, xmin=0.4, xmax=7.1, scale only axis=true,
  x tick label style={font=\scriptsize, inner sep=0.5pt}, xticklabel shift=1.5pt, tick style={draw=none}, major tick length=0pt, y tick label style={font=\scriptsize},
  tick align=outside, xtick style={draw=none}, axis x line*=bottom, axis y line*=left,
  ymajorgrids, grid style={black!12}, every axis title/.style={font=\scriptsize\bfseries, at={(0.5,1)}, anchor=south, align=center},
  area legend, clip=false]
\nextgroupplot[title={MCQ:\\masked slot}, xticklabels={$B$,$A$,$r$,$B$,$A$,$r$}, ylabel={accuracy (\%)}, ylabel style={font=\scriptsize, yshift=-3pt}]
\addplot[fill=none, draw=vizthink, densely dashed, line width=0.5pt, bar shift=-1.6pt, forget plot] coordinates {(1,69.3) (2,75.7) (3,80.0) (4.5,70.7) (5.5,67.7) (6.5,69.7)};
\addplot[fill=none, draw=vizthink, densely dashed, line width=0.5pt, bar shift=1.6pt, forget plot] coordinates {(1,93.0) (2,93.0) (3,99.7) (4.5,88.8) (5.5,85.3) (6.5,96.0)};
\addplot[fill=vizthink, draw=vizthink, bar shift=-1.6pt] coordinates {(1,34.3) (2,71.7) (3,80.0) (4.5,26.2) (5.5,60.3) (6.5,69.7)};
\addplot[fill=white, draw=vizthink, postaction={pattern=north east lines, pattern color=vizthink}, bar shift=1.6pt] coordinates {(1,49.0) (2,91.0) (3,99.7) (4.5,42.9) (5.5,81.7) (6.5,96.0)};
\node[font=\scriptsize\itshape, anchor=north, inner sep=0pt] at ([yshift=-11pt]axis cs:2,0) {Test 1}; \node[font=\scriptsize\itshape, anchor=north, inner sep=0pt] at ([yshift=-11pt]axis cs:5.5,0) {Test 2};
\nextgroupplot[title={MCQ:\\\#Ans}, xticklabels={1,2,3,1,2,3}]
\addplot[fill=none, draw=vizthink, densely dashed, line width=0.5pt, bar shift=-1.6pt, forget plot] coordinates {(1,73.8) (2,77.9) (3,79.1) (4.5,65.9) (5.5,77.1) (6.5,83.7)};
\addplot[fill=none, draw=vizthink, densely dashed, line width=0.5pt, bar shift=1.6pt, forget plot] coordinates {(1,95.7) (2,94.3) (3,93.4) (4.5,89.8) (5.5,89.3) (6.5,93.0)};
\addplot[fill=vizthink, draw=vizthink, bar shift=-1.6pt] coordinates {(1,73.8) (2,32.1) (3,20.9) (4.5,65.9) (5.5,12.9) (6.5,10.5)};
\addplot[fill=white, draw=vizthink, postaction={pattern=north east lines, pattern color=vizthink}, bar shift=1.6pt] coordinates {(1,95.7) (2,33.6) (3,35.2) (4.5,89.8) (5.5,25.0) (6.5,27.9)};
\node[font=\scriptsize\itshape, anchor=north, inner sep=0pt] at ([yshift=-11pt]axis cs:2,0) {Test 1}; \node[font=\scriptsize\itshape, anchor=north, inner sep=0pt] at ([yshift=-11pt]axis cs:5.5,0) {Test 2};
\nextgroupplot[title={OQ:\\conditions $n$}, xticklabels={2,3,4,2,3,4},
  legend to name=strataleg, legend columns=3,
  legend style={font=\scriptsize, draw=black!40, line width=0.4pt, rounded corners=2pt, inner sep=2pt,
    /tikz/every odd column/.append style={column sep=2pt}, /tikz/every even column/.append style={column sep=9pt}}]
\addplot[fill=vizthink, draw=vizthink] coordinates {(1,46.3) (2,46.3) (3,42.4) (4.5,19.6) (5.5,17.1) (6.5,17.3)};
\addplot[fill=white, draw=vizthink, postaction={pattern=north east lines, pattern color=vizthink}] coordinates {(1,68.7) (2,61.2) (3,56.1) (4.5,30.7) (5.5,21.2) (6.5,17.3)};
\addlegendimage{area legend, fill=white, draw=vizthink, densely dashed}
\legend{Llama, Qwen, at least one correct and no wrong}
\node[font=\scriptsize\itshape, anchor=north, inner sep=0pt] at ([yshift=-11pt]axis cs:2,0) {Test 1}; \node[font=\scriptsize\itshape, anchor=north, inner sep=0pt] at ([yshift=-11pt]axis cs:5.5,0) {Test 2};
\end{groupplot}
\end{tikzpicture}
\caption{\ours{} on Test~1 and Test~2 by item construction (accuracy \%; dashed: MCQs scored with at least one correct and no wrong choice).}
\label{fig:strata}
\end{figure}

\begin{table}[t]
\centering
\caption{Sizes of the ontologies (Test~1: source; Test~2: revised).}
\label{tab:onto}
\small
\renewcommand{\arraystretch}{0.9}
\setlength{\tabcolsep}{6pt}
\begin{tabular}{@{}lrrr@{}}
\toprule
 & GO & FoodOn & SNOMED CT \\
\midrule
Named concepts & 51{,}986 & 30{,}995 & 374{,}546 \\
Roles & 9 & 69 & 126 \\
Axioms (Test~1) & 82{,}950 & 44{,}986 & 375{,}261 \\
Axioms (Test~2) & 65{,}810 & 36{,}881 & 375{,}261 \\
\bottomrule
\end{tabular}
\end{table}

\section{{Benchmarks}}
\label{app:data}

\textbf{Table~\ref{tab:onto}} summarizes the sizes of the three ontologies. We use the GO release of 2026-07-26, the FoodOn release of 2022-08-12 (with imports merged), and the SNOMED~CT International Edition release of 2025-07-01, converted to OWL using the SNOMED OWL Toolkit. SNOMED~CT is in the OWL~2 EL profile ($\mathcal{EL}^{++}$). GO uses only $\sqcap$ and $\exists$ in its class expressions and lies in $\mathcal{EL}^{++}$ except for one inverse-object-property axiom relating \emph{part of} and \emph{has part}. FoodOn additionally uses $\forall$, $\neg$, $\sqcup$, and cardinality restrictions, placing it in $\mathcal{SRIQ}$. The revised ontologies used in Test~2 preserve the signature and use a subset of the same constructors.
\textbf{Table~\ref{tab:data}} summarizes the sizes of Tests~1 and~2. For each ontology, we construct 300 MCQs by masking the filler, subject, or relation of 100 axioms each, together with 200 OQs. Test~2 uses the same questions as Test~1, except for 23 questions that fail the checks described below.

\begin{table}[t]
\centering
\caption{Tests~1 (T1) and~2 (T2): MCQs by the number of correct options (\#Ans), and OQs by the number of conditions $n$.}
\label{tab:data}
\renewcommand{\arraystretch}{0.9}%
\small
\setlength{\tabcolsep}{3.2pt}
\begin{tabular}{@{}l@{\hspace{4pt}}l*{6}{>{\raggedleft\arraybackslash}p{16.5pt}}rr@{}}
\toprule
\multicolumn{2}{@{}l}{} & \multicolumn{2}{c}{GO} & \multicolumn{2}{c}{FoodOn} & \multicolumn{2}{c}{SNOMED CT} & \multicolumn{2}{c}{All} \\
\cmidrule(lr){3-4}\cmidrule(lr){5-6}\cmidrule(lr){7-8}\cmidrule(l){9-10}
\multicolumn{2}{@{}l}{} & T1 & T2 & T1 & T2 & T1 & T2 & T1 & T2 \\
\midrule
\multirow{3}{*}[-3pt]{MCQ} & \#Ans${}=1$ & 203 & 203 & 230 & 230 & 236 & 235 & 669 & 668 \\
 & \#Ans${}=2$ & 58 & 58 & 43 & 43 & 39 & 39 & 140 & 140 \\
 & \#Ans${}=3$ & 39 & 39 & 27 & 26 & 25 & 21 & 91 & 86 \\
\midrule
\multirow{3}{*}[-3pt]{OQ} & $n=2$ & 67 & 67 & 67 & 67 & 67 & 65 & 201 & 199 \\
 & $n=3$ & 67 & 67 & 67 & 65 & 67 & 61 & 201 & 193 \\
 & $n=4$ & 66 & 66 & 66 & 62 & 66 & 63 & 198 & 191 \\
\midrule
\multicolumn{2}{@{}l}{All questions} & 500 & 500 & 500 & 493 & 500 & 484 & 1{,}500 & 1{,}477 \\
\bottomrule
\end{tabular}
\end{table}

\textbf{OQ conditions.}\label{app:oq} We construct the candidate OQ conditions as follows. Let \(S(A)\) contain the restrictions \(\exists r.B\) that the MCQ decomposition rule (\S\ref{sec:data-formal}) yields from the asserted axioms of \(A\) and its ancestors, and let the filler be lifted by up to two taxonomy levels:
\[
\overline{S}(A)=\left\{\exists r.B' : \exists r.B\in S(A),\ B'\in\{B\}\cup\mathrm{Parent}_{\leq 2}(B)\right\}.
\]
A lifted restriction still holds for \(A\), since \(B\sqsubseteq B'\). For each \(A\) with \(|\overline{S}(A)|\geq n\), we sample \(n\) conditions from \(\overline{S}(A)\) and resample until none subsumes another (\(\Omc\not\models\exists r_i.B'_i\sqsubseteq\exists r_j.B'_j\) for all \(i\neq j\)).

\textbf{Test~2 checks.}\label{app:t2checks} We apply two checks to ensure that Test~2 remains in the intended pattern (\S\ref{sec:data-hybrid}). First, the revision may remove information needed to preserve the original answer. In particular, replacing an equivalence (\(\bowtie=\equiv\)) with \(A\sqsubseteq D\) drops the reverse implication
\(D\sqcap\bigsqcap_{i}\exists r_i.C_i\sqsubseteq A\),
which is not preserved by the informal annotations. We therefore verify that each question retains the same gold answer when both the revised axioms and the informal annotations are considered. Second, we require that the gold answer cannot be derived from the revised ontology by an ontology reasoner alone.

\section{Construction and Difficulty}
\label{app:strata}

Figure~\ref{fig:strata} shows how the performance of \ours{} varies with item construction, including the number of correct MCQ options, the number of OQ conditions, and the masked slot.

\textbf{Correct options.} For Llama on Test~1, accuracy drops from 73.8\% on single-answer MCQs to 32.1\% and 20.9\% with two and three correct options, respectively, because all entailed options must be identified. Despite this drop, \ours{} remains well above all baselines (at most 14.3\%), as it can verify each option independently with an \code{ASK} query.

\textbf{Conditions.} On Test~1, OQ accuracy is stable up to $n=3$ conditions (46.3\% for Llama at both $n=2$ and $n=3$) and drops to 42.4\% only at $n=4$, since a single \code{SELECT} can evaluate the full conjunction. On Test~2, however, accuracy is far lower throughout and already falls at $n=3$ (from 19.6\% at $n=2$ to 17.1\%), as the required conditions must be recovered from separate definition sentences.

\textbf{Masked slot.} For both models, masking the relation is easiest, while masking the filler is consistently hardest, because the filler's ancestors may also constitute correct answers.

\section{{Prompts}}
\renewtcolorbox{promptbox}[1]{pbox, breakable, title={#1}, before skip=3pt, after skip=3pt, fontupper=\footnotesize, fonttitle=\footnotesize\bfseries, top=1.5pt, bottom=1.5pt, toptitle=0.8pt, bottomtitle=0.8pt}
\label{app:prompts}

\ours{} uses four prompts, reproduced verbatim below: pre-tool reasoning ($\mathcal{P}_{\mathrm{pre}}$), the ReAct loop ($\mathcal{P}_{\mathrm{agent}}$), final answering ($\mathcal{P}_{\mathrm{fin}}$ or $\mathcal{P}_{\mathrm{own}}$, Step~1), and attribution ($\mathcal{P}_{\mathrm{att}}$, Step~2). \textbf{Color code.} Black text is fixed; \slot{this color} marks ontology- or dataset-specific content; and \qslot{red} marks the question block $q$, including the question, answer options, PubMedQA abstract, and answer format. \textbf{Persona.} The persona is prepended to the pre-tool and final-answer prompts. For Tests~1 and~2, SNOMED~CT questions use the persona shown in $\mathcal{P}_{\mathrm{pre}}$, and GO and FoodOn questions replace its first sentence with \emph{``You are an expert in molecular biology''} and \emph{``You are an expert in the science of foods, their ingredients, origins and preparation''}, respectively. For Test~3, we use the dataset-specific OntoTune header.

\begin{promptbox}{Pre-tool reasoning $\mathcal{P}_{\mathrm{pre}}$ (SNOMED~CT persona)}
{\textsc{System:} \slot{You are an expert physician in clinical medicine and healthcare terminology, answering questions about how the entities of that field are defined, organized and related. Answer the following question carefully.} Think step by step briefly to reason about the question. Use the information provided with the question. Additional records are context; use them only when they bear on the question. A draft is unverified. End your reply with the final answer as a single JSON object matching the answer schema, on its own last line.}\\[2pt]
{\textsc{User:} \qslot{\{question block $q$\}}}
\end{promptbox}

\begin{promptbox}{Agent prompt $\mathcal{P}_{\mathrm{agent}}$ (the user turn after the pre-tool reply)}
{An ontology is available through the tool below. Continue from your answer above: where the ontology decides the question, verify that answer with the tools; then give the final answer in the format below.}\\[3pt]
{\textbf{\#\# Tool}}\\
{\code{query(sparql)}: exact reasoning over the ontology's axioms, asked as a short SPARQL query over names: every concept and relation is written as its exact name in double quotes (there are no identifiers or IRIs to use); the relation \code{"is a"} means classification (a kind of). Inheritance and sub-relations are applied for you. The four query forms, each with a toy example:}
\begin{itemize}[leftmargin=1.1em, itemsep=1pt, topsep=1pt, parsep=0pt]
\item {\code{DESCRIBE "name"} -- the ontology's record of one concept: definition, parents, children, the relations it states, annotations. E.g.\ \code{DESCRIBE "grand piano"}}
\item {\code{ASK \{ "subject" "relation" "object" \}} -- is every subject related by relation to some object? Verdicts: ENTAILED, NOT ENTAILED, \mbox{UNKNOWN} NAME. E.g.\ \code{ASK \{ "grand piano" "part of" "orchestra" \}}}
\item {\code{SELECT ?x WHERE \{ ?x "relation" "object" . ... \}} -- the concepts that satisfy ALL the listed conditions; add \code{VALUES ?x \{ "name" "name" ... \}} to test only these candidates, one verdict each. E.g.\ \code{SELECT ?x WHERE \{ ?x "part of" "orchestra" .\ ?x "made of" "brass" \}}}
\item {\code{SELECT ?o WHERE \{ "subject" "relation" ?o \}} -- what the subject is related to by that relation; likewise \code{SELECT ?r WHERE \{ "subject" ?r "object" \}} lists the relations that hold between the two. E.g.\ \code{SELECT ?o WHERE \{ "grand piano" "part of" ?o \}}}
\end{itemize}
{\textbf{\#\# How to use it}}\\
{Use the ontology when the question is about how its concepts are defined, classified or related to each other: query decides such a question exactly, so verify the candidate answers with it before answering. Answer a question about a situation, a case, a measurement or general knowledge from your own knowledge. Names must be spelled exactly as the ontology spells them; if a name comes back \mbox{UNKNOWN} NAME, retry with one of the similar names it lists.}\\[3pt]
{\textbf{\#\# What the verdicts mean}}\\
{ENTAILED is certain: the ontology's axioms derive it. NOT ENTAILED is not a denial: the ontology is incomplete (open world), so a statement it does not derive may still hold. Decide such a statement yourself from the records shown and from what you know, and accept it when it is clearly true.}\\[3pt]
{\textbf{\#\# Format}}\\
{\code{Thought:} reason about the question and about what to do next -- take as much room as you need}\\
{\code{Action:} one tool call, written as a function call: \code{query(<SPARQL query>)}}\\
{\code{Observation:} the tool's result, given to you -- never write one yourself}\\
{\ldots\ (this repeats as often as needed)}\\
{\code{Thought:} I now know the answer}\\
{\code{Answer:} <ONLY the JSON object in the answer schema the question states>}\\[2pt]
{Write a Thought before every Action and before the Answer. A Thought may also stand on its own; the reply after it must be an Action or an Answer. Begin.}
\end{promptbox}

\begin{promptbox}{Final answer: $\mathcal{P}_{\mathrm{fin}}$ (non-empty digest) and $\mathcal{P}_{\mathrm{own}}$ (empty digest)}
{$\mathcal{P}_{\mathrm{fin}}$, \textsc{System:} The tool phase of this session is over. Using the question and the observations below, think step by step briefly, then end your reply with the final answer JSON object that the question's answer schema requests, on its own last line. ENTAILED is certain: the ontology's axioms derive it. NOT ENTAILED is not a denial: the ontology is incomplete (open world), so a statement it does not derive may still hold. Decide such a statement yourself from the records shown and from what you know, and accept it when it is clearly true.}\\[2pt]
{$\mathcal{P}_{\mathrm{fin}}$, \textsc{User:} \slot{\{persona\}} \qslot{\{question block $q$\}}}\\
{Observations so far:}\\
{\slot{\{the digest $E$\}}}\\
{Draft answer from the tool phase: \slot{\{draft $\tilde y$\}}}\\[4pt]
{$\mathcal{P}_{\mathrm{own}}$, \textsc{System:} Answer the question from your own knowledge. Think step by step briefly, then end your reply with the final answer JSON on its own last line.}\\[2pt]
{$\mathcal{P}_{\mathrm{own}}$, \textsc{User:} \slot{\{persona\}} \qslot{\{question block $q$\}}}
\end{promptbox}

\begin{promptbox}{Attribution $\mathcal{P}_{\mathrm{att}}$ (one more user turn after the reply of $\mathcal{P}_{\mathrm{fin}}$; \S\ref{sec:grounding})}
{Your answer above is final; do not change it. Now state what it rests on. The observations you were given, numbered:}\\[2pt]
{\slot{[E1] \{observation $e_1$\}}}\\
{\slot{[E2] \{observation $e_2$\} \ldots}}\\[2pt]
{Reply with only this JSON object:}\\
{\code{\{"evidence": [<the ids of the observations your answer relies on, e.g. "E1", "E3"; empty if none>], "grounding": "<one word: entailed, supported, related or unrelated>"\}}}\\
{Choose the grounding as the first of these that holds:}
\begin{itemize}[leftmargin=1.1em, itemsep=1pt, topsep=1pt, parsep=0pt]
\item {entailed: your answer follows from the observations by strict logical reasoning: for each option or name in your answer, the facts stated in the observations derive it, with nothing added from your own knowledge. A fact counts whether the tool states it formally (an ENTAILED verdict, a listing of the concepts entailed to satisfy the question's conditions, a stated relation, parent or child) or a text field states it as explicitly as an axiom would, naming the concepts and the relation (e.g.\ a definition sentence ``Every X is related by R to some Y''). Concept names, synonyms and keyword records marked ``not verified'' state no fact.}
\item {supported: your answer follows from the observations by sound reasoning that is not such a derivation: it interprets text that does not state the fact explicitly (a paraphrasing definition, a name or synonym match, a ``not verified'' keyword record), or it combines the facts shown with your own knowledge.}
\item {related: the observations give some valid information relevant to the question, but not enough to arrive at your answer, or they point to a different answer.}
\item {unrelated: the observations are essentially irrelevant to the question and to your answer.}
\end{itemize}
{NOT ENTAILED or ``0 concepts'' only means that the axioms do not derive a statement (open world); it does not deny it. So an empty answer (no option selected, an empty list), which claims that nothing satisfies the question, is at most related.}\\[2pt]
{If the reply holds no such object with one of the four levels, one re-ask follows: \emph{Reply with only the JSON object \code{\{"evidence": [...], "grounding": "..."\}}. The first character of the reply must be `\{' and the last must be `\}'.}}
\end{promptbox}

\clearpage
\end{document}